\documentclass{article}

\usepackage{arxiv}

\usepackage[utf8]{inputenc} 
\usepackage[T1]{fontenc}    
\usepackage{hyperref}       
\usepackage{url}            
\usepackage{booktabs}       
\usepackage{amsfonts}       
\usepackage{nicefrac}       
\usepackage{microtype}      
\usepackage{lipsum}
\usepackage{graphicx}
\usepackage{url,hyperref,lineno,microtype,subcaption}
\usepackage{algorithm}
\usepackage{algpseudocode}
\usepackage{float}
\usepackage{amsmath,amssymb,amsthm}
\usepackage{mathtools}
\usepackage{wrapfig}
\graphicspath{ {./images/} }

\title{Learning to learn online with neuromodulated synaptic plasticity in spiking neural networks}

\author{
 Samuel Schmidgall \\
  U.S. Naval Research Laboratory \\
  \and \textbf{Joe Hays} \\
 U.S. Naval Research Laboratory \\
}

\begin{document}
\maketitle
\begin{abstract}
We propose that in order to harness our understanding of neuroscience toward machine learning, we must first have powerful tools for training brain-like models of learning. Although substantial progress has been made toward understanding the dynamics of learning in the brain, neuroscience-derived models of learning have yet to demonstrate the same performance capabilities as methods in deep learning such as gradient descent. Inspired by the successes of machine learning using gradient descent, we demonstrate that models of neuromodulated synaptic plasticity from neuroscience can be trained in Spiking Neural Networks (SNNs) with a framework of learning to learn through gradient descent to address challenging \textit{online} learning problems. This framework opens a new path toward developing neuroscience inspired online learning algorithms.
\end{abstract}


\section*{Introduction}

The ability to learn continually across vast time spans is a hallmark of the brain which is unrivaled by modern machine learning algorithms. Extensive research on learning in the brain has provided detailed models of synaptic plasticity--however, these models of learning have yet to produce the impressive capabilities demonstrated by deep neural networks trained with backpropagation. Despite our increasing understanding of biological learning, the most powerful methods for optimizing neural networks have remained backpropagation-based. However, when backpropagation is applied to a continuous stream of data, issues arise since gradient descent approaches do not address the ability to update synapses continually without forgetting previously learned information. This is because backpropagation methods modify the weight of every synapse at every update, which causes task-specific information from previous updates to rapidly deteriorate. The tendency for backpropagation to overwrite previously learned tasks has made its use as an online learning algorithm impractical \cite{doi:10.1073/pnas.1611835114, PARISI201954}. The brain solves this problem by determining its own modification as a function of information that is locally available to neurons and synapses. This ability for self-modification is a process that has been fine-tuned through a long course of evolution, and is the basis of learning and memory in the brain \cite{evobones}. Can the success of gradient descent be combined with neuroscience models of learning in the brain?

Recent experimental evidence from neuroscience has provided valuable insight into the dynamics of learning in the brain \cite{fremaux2016neuromodulated, gerstner2018eligibility}. Two fundamental findings have led to recent successes in the development of online neuro-inspired learning algorithms \cite{bellec2020solution, Schmidgall2021SpikePropamineDP, 10.3389/fnbot.2019.00081, doi:10.1073/pnas.2111821118, KUSMIERZ2017170}. First, neurons and synapses in the brain maintain historical traces of activity. These traces, referred to as eligibility traces, are thought to accumulate the joint interaction between pre- and post-synaptic neuron factors. Eligibility traces do not automatically produce a synaptic change, but have been demonstrated to induce synaptic plasticity in the presence of top-down learning signal. Second, the brain has a significant quantity of top-down learning signals which are broadly projected by neurons from higher centers in the brain to plastic synapses to convey information such as novelty, reward, and surprise. These top-down signals often represent neuromodulator activity such as dopamine \cite{steinberg2013causal, schultz1993responses, seamans2007dopamine, zhang2009gain, Speranza2021DopamineTN} or acetylcholine \cite{ranganath2003neural, teles2013cholinergic, brzosko2015retroactive, hasselmo2006role, zannone2018acetylcholine, Zannone2018AcetylcholinemodulatedPI, Hasselmo2006TheRO}. The interaction between eligibility traces and top-down learning signals enables learning rules to connect interactions between long and short time scales \cite{fremaux2016neuromodulated, gerstner2018eligibility}.


Here, we demonstrate that models of neuromodulated synaptic plasticity from neuroscience can be trained in SNNs with the paradigm of learning to learn through gradient descent. These results demonstrate that neuromodulated synaptic plasticity rules can be optimized to solve temporal learning problems from a continuous stream of data, leading to dynamics that are optimized to address several fundamental \textit{online} learning challenges. This new paradigm allows models of neuromodulated synaptic plasticity to realize the benefits from the success of gradient descent in machine learning while staying true to neuroscience. This opens the door for validating learning theories in neuroscience on challenging problems, as well as developing effective online learning algorithms which are compatible with existing neuromorphic hardware.















\section*{Learning in networks with plastic synapses}


\textbf{Learning how to learn online.} The primary strategy for developing online learning systems has been to attempt discovering each piece of the system manually such that these pieces can one day be assembled to form an effective online learning system. Alternatively, the paradigm of meta-learning aims to learn the learning algorithm itself such that it ultimately discovers a solution that solves the inherent learning problems out of necessity \cite{clune2019ai}. Meta-learning has been notoriously difficult to define, and is often used inconsistently across experiments--however, it is consistently understood to signify \textit{learning how to learn}: improving the learning algorithm itself \cite{hospedales2020meta}. More concisely, meta-learning is a learning paradigm that uses meta-knowledge from previous experience to improve its ability to learn in new contexts. This differs from \textit{multi-task} learning in that, multi-task learning aims to produce a model that performs well on multiple tasks that are explicitly encountered during the optimization period, whereas meta-learning primarily aims to produce a model that is able to learn novel tasks more efficiently. 

Meta-learning consists of an inner (base) and outer (meta) loop learning paradigm \cite{hospedales2020meta}. During \textbf{base learning}, an inner-loop learning algorithm solves a task, such as robotic locomotion or image classification, while optimizing a provided objective. During \textbf{meta-learning}, an outer-loop (\textit{meta}) algorithm uses information collected from the base learning phase to improve the inner-loop (base learning) algorithm toward optimizing the outer-loop objective. It is proposed that there are three axes within the meta-learning paradigm: meta-representation (what?), meta-optimization (how?), and meta-objective (why?) \cite{hospedales2020meta}. The meta-representation refers to the representation of meta-knowledge $\omega$. This knowledge could be anything from initial model parameters \cite{finn2017model, rothfuss2018promp, fakoor2019meta, liu2019taming}, the inner optimization process \cite{metz2018meta, andrychowicz2016learning, irie2022modern, bello2017neural}, or the model architecture \cite{zoph2016neural, real2019regularized, lian2019towards, liu2018darts}. The meta-optimizer refers to the choice of optimization for the outer-level in the meta-training phase which updates meta-knowledge $\omega$. The meta-optimizer often takes the form of gradient-descent \cite{finn2017model}, evolutionary strategies \cite{houthooft2018evolved}, or genetic algorithms \cite{co2021evolving}. The meta-objective specifies the goal of the outer-loop learning process, which is characterized by an objective $\mathcal{L}^{meta}$ and task distribution $\mathcal{D}^{test (i)}_{source}$.


To provide a more formal definition, the bilevel optimization perspective of meta-learning is presented as follows:

\begin{equation}\label{eq:Meta}
\boldsymbol \omega^{*} = \underset{\boldsymbol \omega}{\text{arg min}} \  \sum_{i=1}^{M} \mathcal{L}^{meta} (\theta^{* (i)}(\omega), \omega, \mathcal{D}^{test (i)}_{source})
\end{equation}

\begin{equation}\label{eq:MetaCond}
\text{s.t.  } \theta^{* (i)}(\omega) = \underset{\boldsymbol \theta}{\text{arg min }} \mathcal{L}^{task}(\theta, \omega,  \mathcal{D}^{train (i)}_{source}).
\end{equation}

Equation \ref{eq:Meta} represents the outer-loop optimization, which looks to find an optimal meta-representation $\omega^*$ defined by the selection of values $\omega$ such that the meta-objective loss $\mathcal{L}^{meta}$ is minimized across a set of $M$ tasks from the task testing distribution $\mathcal{D}^{test (i)}_{source}$. The minimization of $\mathcal{L}^{meta}$ is dependent on finding $\theta^{* (i)}(\omega)$, which is the selection of values for $\theta$ that minimize the task loss $\mathcal{L}^{task}$ using the meta-representation $\omega$. In other words, $\theta^{* (i)}(\omega)$ looks to finds the optimal $\theta$ for a given \textit{training} distribution of data using $\omega$ and $\omega^*$ looks to find the optimal $\omega$ for a given \textit{testing} distribution with $\theta^{* (i)}(\omega)$ that was optimized on the training distribution using a given $\omega$.


\begin{figure*}
    \centering
    \includegraphics[width=0.99\linewidth]{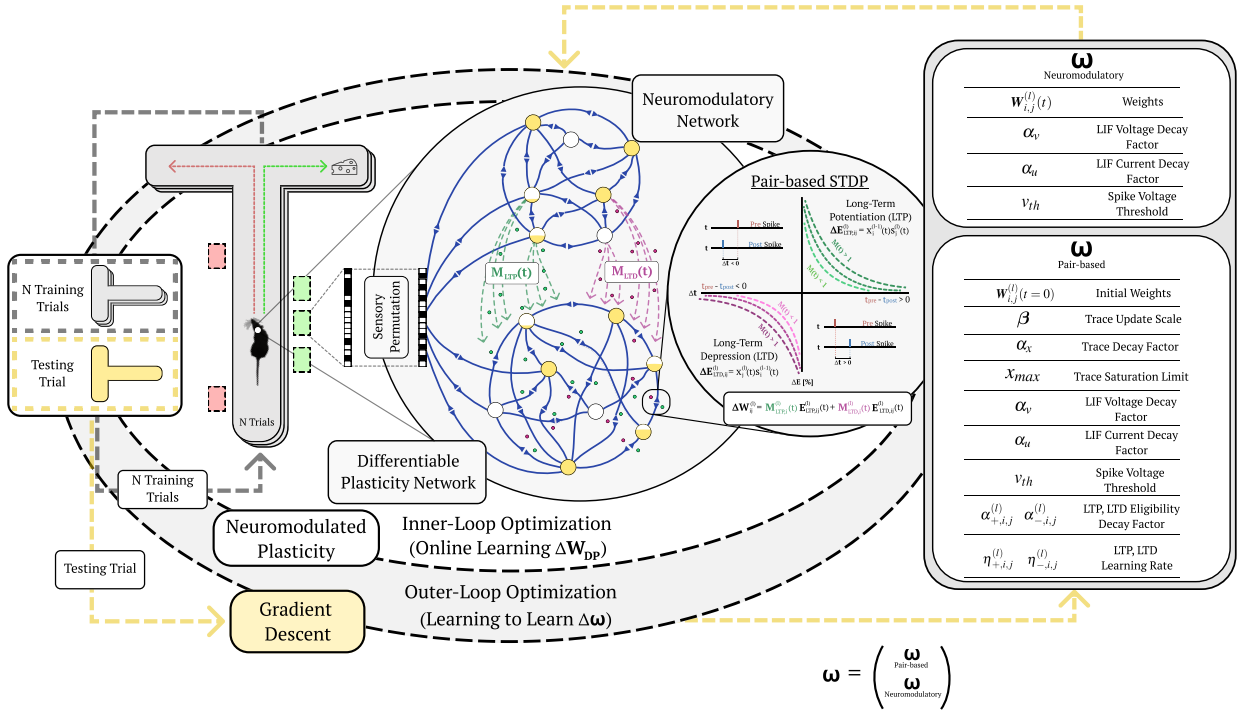}
    \caption{\textbf{Learning to learn online with neuromodulated synaptic plasticity.} An example of the meta-learning paradigm on a one-shot cue association problem. A virtual rodent travels down a T-maze for a series of trials with a novel randomly permuted sensory input, and must learn the representation of the novel permutation through the inner-loop optimization via synaptic plasticity (pair-based STDP) and neuromodulation from N training trials before it is evaluated on a testing trial. The outer-loop representation optimizes (through gradient descent) the plasticity and neurmodulatory parameters ($\omega$) to better learn from novel random permutations during the training trials; meaning, the (inner) base learning, which optimizes Equation \ref{eq:MetaCond}, is accomplished through the network dynamics learned by the (outer) meta learning Equation \ref{eq:Meta} (which is solved by gradient descent). This illustrates learning to learn, where Equation \ref{eq:Meta} is learning how to make the network learn (i.e. solve Equation \ref{eq:MetaCond}). Permuted sensory cues are sent to the Differentiable Plasticity SNN (DP-SNN), which has plastic synapses, and the Neuromodulatory SNN (NM-SNN), which sends top-down signals that modulate plastic synapses in the DP-SNN.}
    \label{figure:RodentVisual}
\end{figure*}


\textbf{Learning how to learn online with synaptic plasticity through gradient descent.} In learning applications with networks of spiking neurons, synaptic plasticity rules have historically been optimized through black-box optimization techniques such as evolutionary strategies \cite{bohnstingl2019neuromorphic, Schmidgall2020AdaptiveRL, schmidgall2021stable}, genetic algorithms \cite{ jordan2021evolving, elbrecht2020neuroevolution}, or Bayesian optimization \cite{kulkarni2021training,nessler2008hebbian}. This is because spiking dynamics are inherently non-differentiable, and non-differentiable computations prevent gradient descent from being harnessed for optimization. However, recent advances have developed methods for backpropagating through the non-differentiable part of the neuron with surrogate gradients \cite{shrestha2018slayer,neftci2019surrogate,zenke2021remarkable}, which are continuous relaxations of the true gradient. These advances have also allowed gradient descent based approaches to be utilized for optimizing both the parameters defining plasticity rules and neuromodulatory learning rules in SNNs \cite{Schmidgall2021SpikePropamineDP}. However, previous work optimizing these rules use neuromodulated plasticity as a dynamic which \textit{compliments} the network on tasks which can be solved without it instead of using it as the learning algorithm itself \cite{Schmidgall2021SpikePropamineDP}. Methods which \textit{do} use neuromodulated plasticity as a learning algorithm do not learn its dynamics from biological learning rules, but define rules which are derived from machine learning approaches \cite{scherr2020one, bellec2020solution}. Instead, we desire to provide a paradigm of using learning rules from neuroscience that can be optimized to act as the learning algorithm through gradient descent.

An insight which enables this is the idea that synaptic plasticity in the presence of a neuromodulatory signal can be thought of as a meta-learning optimization process, with meta-knowledge $\omega$ being represented by the learned plasticity rule parameters and $\theta$ as the strength of synaptic weights representing the inner-level free parameters which change based on $\omega$ \textit{online}. Since both the parameters governing the dynamics of the neuromodulatory signal and the plasticity rules in SNNs can be optimized through backpropagation through time (BPTT) \cite{Schmidgall2021SpikePropamineDP}, the outer-loop training can be framed to optimize neuromodulated plasticity rules (Equation \ref{eq:Meta}) which act as the inner-loop learning process (Equation \ref{eq:MetaCond}). The optimization goal of outer-loop in Equation \ref{eq:Meta} is a selection of the neuromodulatory and plasticity parameters for $\omega$ which minimize the outer-loop loss $\mathcal{L}^{meta}$ as a function of $\theta^{* (i)}(\omega)$, $\omega$, and $\mathcal{D}^{test (i)}_{source}$. The optimization goal of the inner loop in Equation \ref{eq:MetaCond} is to find $\theta^{* (i)}(\omega)$ which is defined as a selection of the parameters for $\theta$ which minimize the inner-loop loss $\mathcal{L}^{task}$, such that $\theta$ is determined across time as a function of the plasticity equation and $\omega$, which parameterizes the plasticity rules and the neuromodulatory dynamics, for a given task $\mathcal{D}^{train (i)}_{source}$. By optimizing the learning process, gradient descent, which acts on $\omega$ in Equation \ref{eq:Meta}, is able to shape the dynamics of learning in Equation \ref{eq:MetaCond} such that it is able to solve problems that gradient descent is not able to solve on its own. To do this, learning problems are presented to emulate how biological organisms are trained to solve tasks in behavioral experiments--specifically with respect to the \textit{online} nature of the task. The meta-learning process can then shape plasticity and neuromodulatory dynamics to address more fundamental challenges that are presented during the inner-loop task. Rather than manual design, these fundamental learning problems are addressed implicitly by the optimization process out of necessity for solving the meta-learning objective. As this work will demonstrate, using neuromodulated plasticity as the meta-representation allows for the learning algorithm itself to be learned, making this optimization paradigm capable of \textit{learning} to solving difficult temporal learning problems. This capability is demonstrated on an online one-shot continual learning problem and on a online one-shot image class recognition problem.

\section*{Experiments}

\subsection*{One-shot continual learning: Addressing credit assignment through one-shot cue association}



Experience-dependent changes at the synapse serve as a fundamental mechanism for both short- and long-term memory in the brain. These changes must be capable of attributing the outcome of behaviors together with the necessary information contained in temporally-dependent sensory stimuli, all while ignoring irrelevant details; if the behavior produced by a particular stimuli led to a good outcome it should be reinforced and visa versa. The problem, however, is that the outcome of behavior is often not realized for a long and typically variable amount of time after the actions affecting that outcome are produced. Additionally, there are often many elements of sensory noise that could serve to distract the temporal-learner from proper credit assignment. 

To examine these capabilities, a valuable learning experiment from neuroscience tests the cognitive capabilities of rodents in a T-maze. The T-maze can be described as an enclosed structure that takes the form of a horizontally-placed T \cite{lett1975long, wenk1998assessment, dudchenko2001animals, deacon2006t, engelhard2019specialized}, with the maze beginning at the base of T and ending at either side of the arms, see Figure \ref{figure:RodentVisual}. The rodent moves down the base of the maze and chooses either side of the arms. In some experiments, a series of sensory cues are arranged along the left and right of the apparatus as the rodent makes progress toward the end of the maze. A decision as to which side of the maze will provide positive and negative reinforcement is based on the arrangement of these stimuli \cite{morcos2016history, engelhard2019specialized}. The rodent is rewarded for choosing the side of the track with the majority of cues. This task is not trivial to solve since the rodent has to recognize that the outcome is not effected by the presentation ordering of the cues or which side the last cue was on. Rather, the cues must be counted independent of their ordering for each side and the sums must be compared to make a decision. Making learning more difficult, the reward for solving this problem is not presented until after a decision has been made, so the rodent must address credit assignment for its behavior across the time span of an entire cue experiment.

\begin{figure*}
    \centering
    \includegraphics[width=0.93\linewidth]{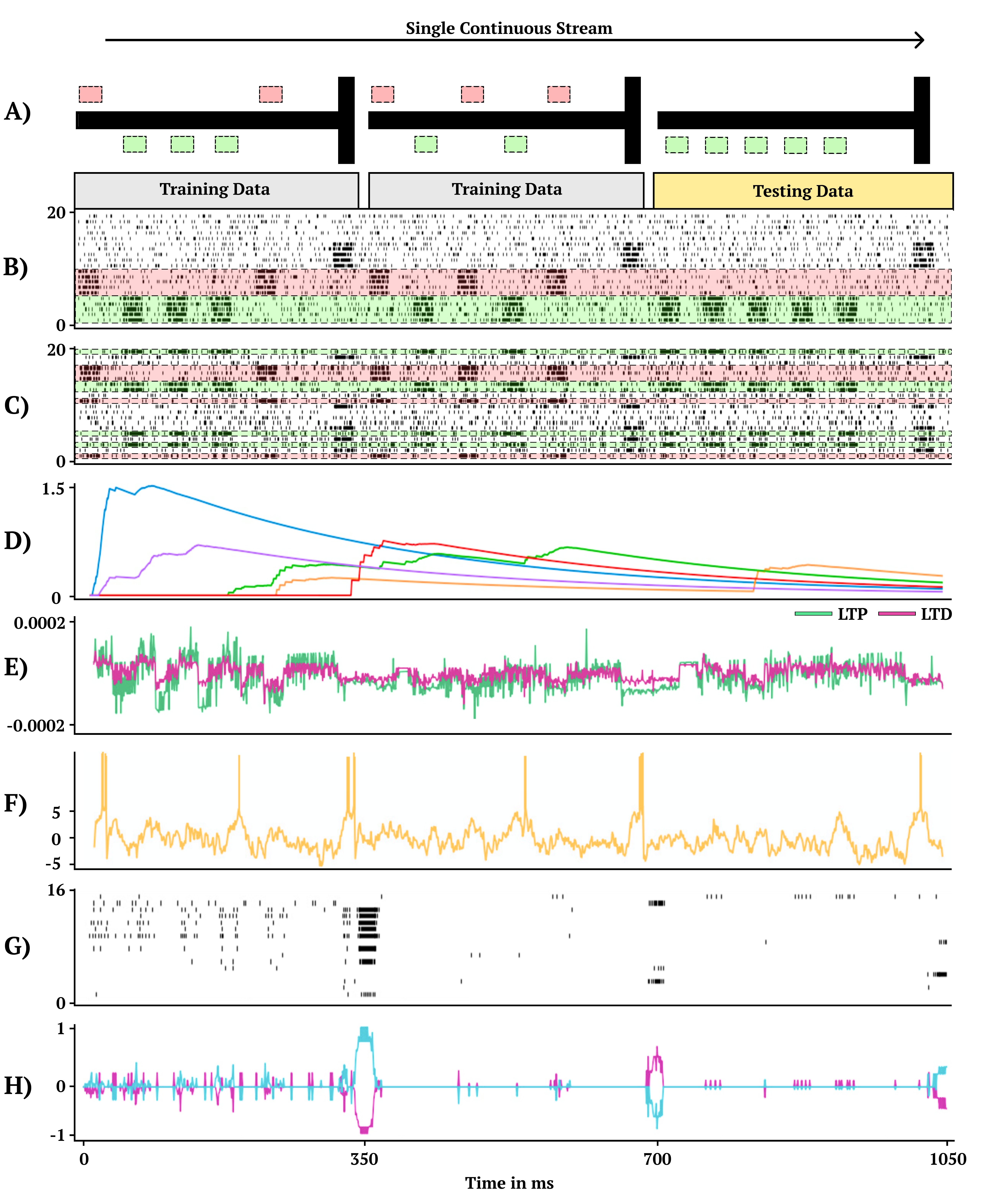}
    \caption{\textbf{One-shot cue association.} Visual demonstration of the continual one-shot learning paradigm for a trained neuromodulatory network. (A) Graphical interpretation of cue-association task. Two training data mazes are presented in a random order, one from each class of right (first maze) and left (second maze) cues followed by a testing data maze. Gray corresponds to training data which receives no reward to backpropagate and gold corresponds to testing data which does receive reward to backpropagate. (B) Non-permuted sensory information represented as spikes indexed from 0 to 20 (bottom to top). (C) Permuted form of sensory information presented in B indexed by which neuron receives spikes. (D) Eligibility trace dynamics (Methods, Equation \ref{eq:Eligibility}) sampled from five random synapses. (E) LTP (green) and LTD (magenta) neuromodulatory dynamics from a random modulatory neuron. (F) Activity of a hidden neuron. (G) Sample of 16 hidden neuron spiking activity. (H) Action neuron activity.}
    \label{figure:CueAssociation}
\end{figure*}

Previous experiments with SNNs in simulation have demonstrated that synaptic plasticity alone enables a network to solve this problem where it was not able to be solved with feedforward SNNs \cite{Schmidgall2021SpikePropamineDP} or Recurrent SNNs (RSNNs) \cite{bellec2020solution} using BPTT. However, previous work only considered learning in this environment in a setting where the neurons associated with a particular cue remained consistent across gradient updates and experiments. In this way, there was no inner- and outer-level optimization. Rather, the synaptic plasticity served as a mechanism for memorization and cue-decision making, but not actually learning which cues and which decisions are associated with positive reward during the network time-horizon, and hence it does not qualify as a meta-learning problem. Additionally, during \textit{in-vivo} rodent experiments, accurate cue-problem performance is demonstrated with only 7-12 sessions per mouse \cite{morcos2016history}. This differs from the learning efficiency of ref. \cite{Schmidgall2021SpikePropamineDP} and ref. \cite{bellec2020solution}, which take on the order of hundreds and thousands of training sessions respectively.


\textbf{Many- and one-shot learning.} Converting this experiment from neuroscience into simulation, sensory cues are emulated as probabilistic spike trains, with subgroups of neurons corresponding to particular sensory cues. Twenty sensory neurons are organized into four subgroups, five of which represent right-sided cues, five for left-sided cues, five of which display activity during the decision period, and five which purely produce spike noise (Figure \ref{figure:CueAssociation}B). To transform this problem into a meta-learning problem, the particular sensory neurons which are associated with cues, decision timings, and noise are randomly \textit{permuted} (Figure \ref{figure:CueAssociation}C) making the temporal learner unable to know which neurons are associated with which stimuli at the beginning of each cue-association task. The network is then presented with a series of cue-trials (Figure \ref{figure:CueAssociation}A) and a reward signal at the end of each trial. The many-shot cue association experiment is as follows: (1) the neurons associated with particular cues in previous experiments are randomly permuted, (2) the network is placed at the beginning of the cue-maze, (3) a series of sensory inputs, noise, cues, and decision activity are input into the sensory neurons as the learner moves along the apparatus, (4) at the end of the maze the learner makes a decision (left or right) based on the sensory input and a reward signal is provided as input to the neuromodulatory network based on whether it was the correct decision, (5) the agent is placed at the beginning of the maze and starts from step 2 \textit{without} resetting network parameters and traces for $N$ trials for all $K$ cues (left and right) acting as a \textit{training} phase (i.e. inner loop solving Equation \ref{eq:MetaCond}), (6) the performance of the network is \textit{tested} based on the information that has been learned from the $N$-shot cue presentations, (7) plasticity parameters are updated through gradient descent based on evaluation performance for the final test trial (i.e. outer loop solving Equation \ref{eq:Meta}), and the learning problem is repeated from step 1. The benefit of permuting the sensory neurons as a source of inner-loop learning is that it results in a large number of variations of the problem. With only 20 neurons there are $20! = 2.4\cdot10^{18}$ variations, which results in learning experiments which are unlikely to have repetitions in the problem domain.

One-shot learning is a particularly challenging variation of the $N$-shot learning paradigm, where $N$ is set equal to one for each of the $K$ classes. In this way, the learning model is only provided with \textit{one} example of each class and must be capable of differentiating between classes based only on the given single example. One-shot learning is argued to be one of two important capabilities of the brain that is missing from models of learning in computational neuroscience \cite{BREA201661}.


\begin{wrapfigure}{R}{0.3\textwidth}
    \begin{center}
        \includegraphics[width=0.28\textwidth]{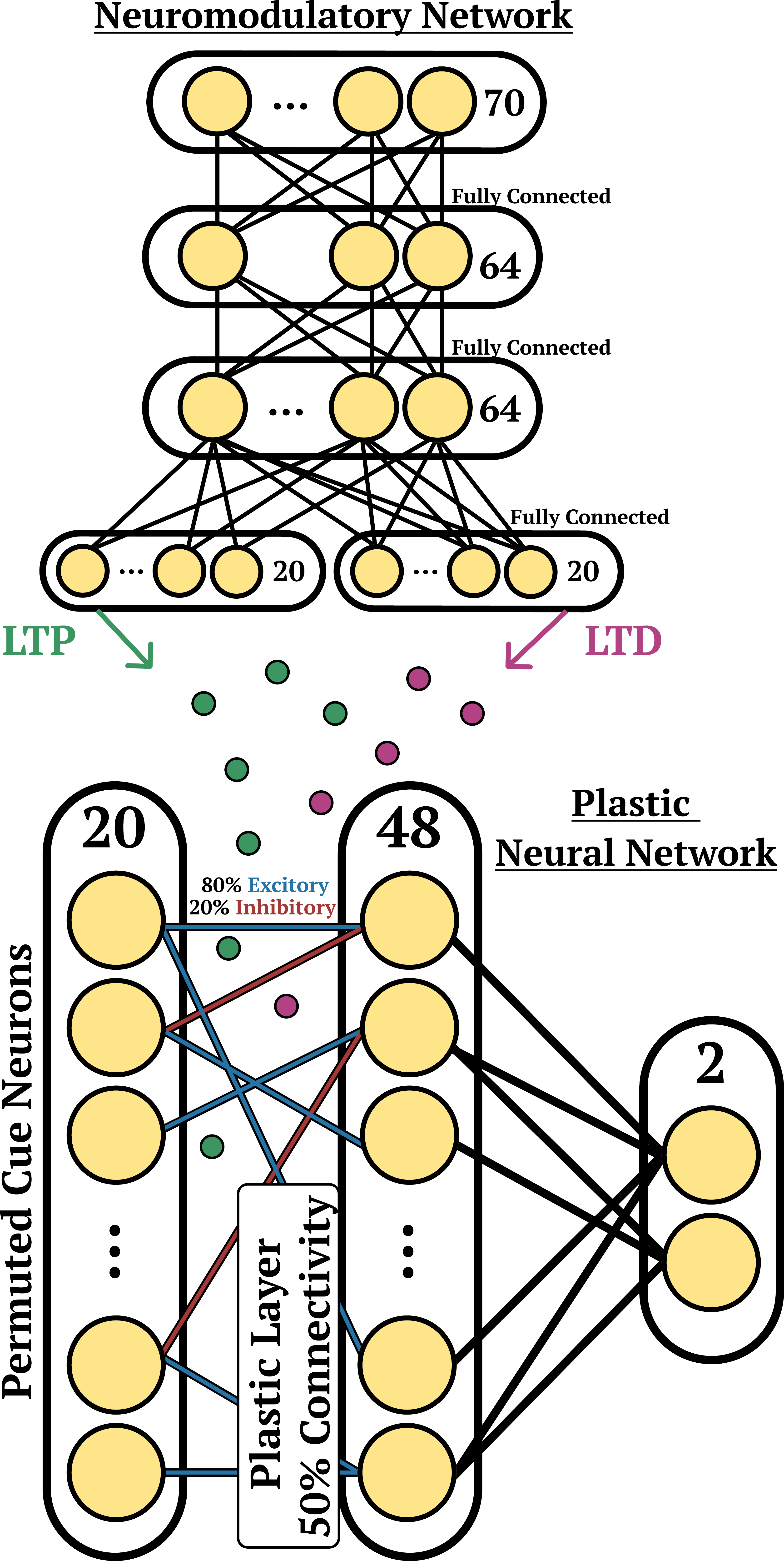}
    \end{center}
    \caption{\textbf{Cue Association Architecture.} Depiction of the network structure for the DP-SNN (bottom) and the NM-SNN (top) for the cue association experiment.}
    \label{figure:NetStructCue}
\end{wrapfigure}

\textbf{Architecture.} The DP-SNN in Figure \ref{figure:NetStructCue} contains one-hidden layer of 48 Current-Based Leaky Integrate-and-Fire (CUBA) neurons (Methods, Equation \ref{eq:CurrentCUBA}-\ref{eq:VoltageCUBA}). Synaptic connections between the input neurons and the hidden layer neurons accumulate changes in an eligibility trace based on an additive pair-based STDP rule (Methods, Equation \ref{eq:STDPAdd}) with a Long-Term Potentiation (LTP) trace for the LTP dynamics of the pair-based rule and an Long-Term Depression (LTD) trace for the LTD dynamics. Pair-based STDP (Methods, Equation \ref{eq:PairbasedSTDP}) represents plasticity based on the product of timing relationships between pairs of pre- and post-synaptic activity. Learning signals for the LTP trace and the LTD trace are produced by an independent neuromodulatory SNN (NM-SNN) using an input neuron specific modulatory signal (Methods, Equation \ref{eq:ModEligibilityNeuronMod}) for both the LTP and LTD dynamics. Connections from the hidden neurons to the output activity are non-plastic synapses learned through gradient descent. During network initialization, only a fraction of neurons are connected, with a connection probability of $50\%$. Each initialized synapse is assigned to represent either an inhibitory synapse with $20\%$ probability or an excitatory synapse with $80\%$ probability, with inhibitory synapses producing negative currents in outgoing neurons and excitatory synapses producing positive ones. The neuromodulatory SNN contains two layers of 64 CUBA neurons (Methods, Equation \ref{eq:CurrentCUBA}-\ref{eq:VoltageCUBA}). The synapses are \textit{non-plastic} and are fully-connected between layers.  The NM-SNN receives the same sensory input as the DP-SNN in addition to the DP-SNN hidden neuron activity and a learning signal that occurs at the decision interval for the training cue sequences. Both the DP-SNN and the neuromodulatory SNN share the same meta-objective and are optimized jointly in an end-to-end manner with \textit{BPTT} in the outer loop (i.e. Equation \ref{eq:Meta}). Error for the one-shot learning task is calculated via binary cross entropy loss on the output neuron activity compared with the correct cue label (Figure \ref{figure:CueAssociation}H) during the testing data trajectory, with $\mathcal{L}^{meta} = -\sum_{i=1}^{2}(\text{log}(p(y_{i})) + (1-y_{i})\text{log})(1-p(y_{i}))$ and $y_{i}$ equal to the weighted output neuron activity.

\textbf{Experimental setup.} The one-shot learning experiment in this work presents $M=5$ cues (Figure \ref{figure:CueAssociation}A-C). During a cue presentation period the permuted cue neuron has a firing probability of $0.75$. When the cue neuron is not active (during a cue presentation) the firing probability is $0.15$. The noise neurons have a firing probability of $0.15$ at each moment in time and the decision interval neurons have a firing probability of $0.75$ during a decision period and $0.15$ otherwise. The cue presentation period for each cue spans 25 ms which is followed by a 30 ms resting period between each cue. After the final cue there is a 50 ms resting period before the decision period which is 25 ms totalling 350 ms for each individual cue problem in the one-shot cue-association task. The simulation step size is set to 1 ms. An environment feedback signal arrives at the end of each cue-trial during the decision interval, requiring the synapses to store and process the necessary information relating the permuted input cues and the learning signal. This signal is only given to the neuromodulatory network during the training data phase (Figure \ref{figure:CueAssociation}A). The environment learning signal is two-dimensional binary vector, with the first element as one during a right-cue task, the second element as one during a left cue-task, and each element is otherwise zero.

\textbf{Results.} Synaptic plasticity occurs continuously at every moment in time rather than during select periods. Task specific knowledge is not able to be transferred between cue streams since cue frequency, cue ordering, and input permutations are randomly ordered. Rather information must be transferred between cue streams by improving online learning via the optimization of the meta-representation of plasticity--improving the learning algorithm itself (i.e. solving Equation \ref{eq:Meta}). Recalling the definition of continual learning from ref. \cite{delange2021continual}, information within a cue stream must be retained and improved upon across the two presented training trials without clear task divisions being provided. Unimportant information in the form of noise neurons and random cue firings must be selectively recognized and forgotten. Critically, this requires the optimized learning algorithm to store learned information in synapses from the training cue trials without catastrophically forgetting in order to solve the testing cue trial.

\begin{wrapfigure}[20]{L}{0.40\textwidth}
    \begin{center}
        \includegraphics[width=0.37\textwidth]{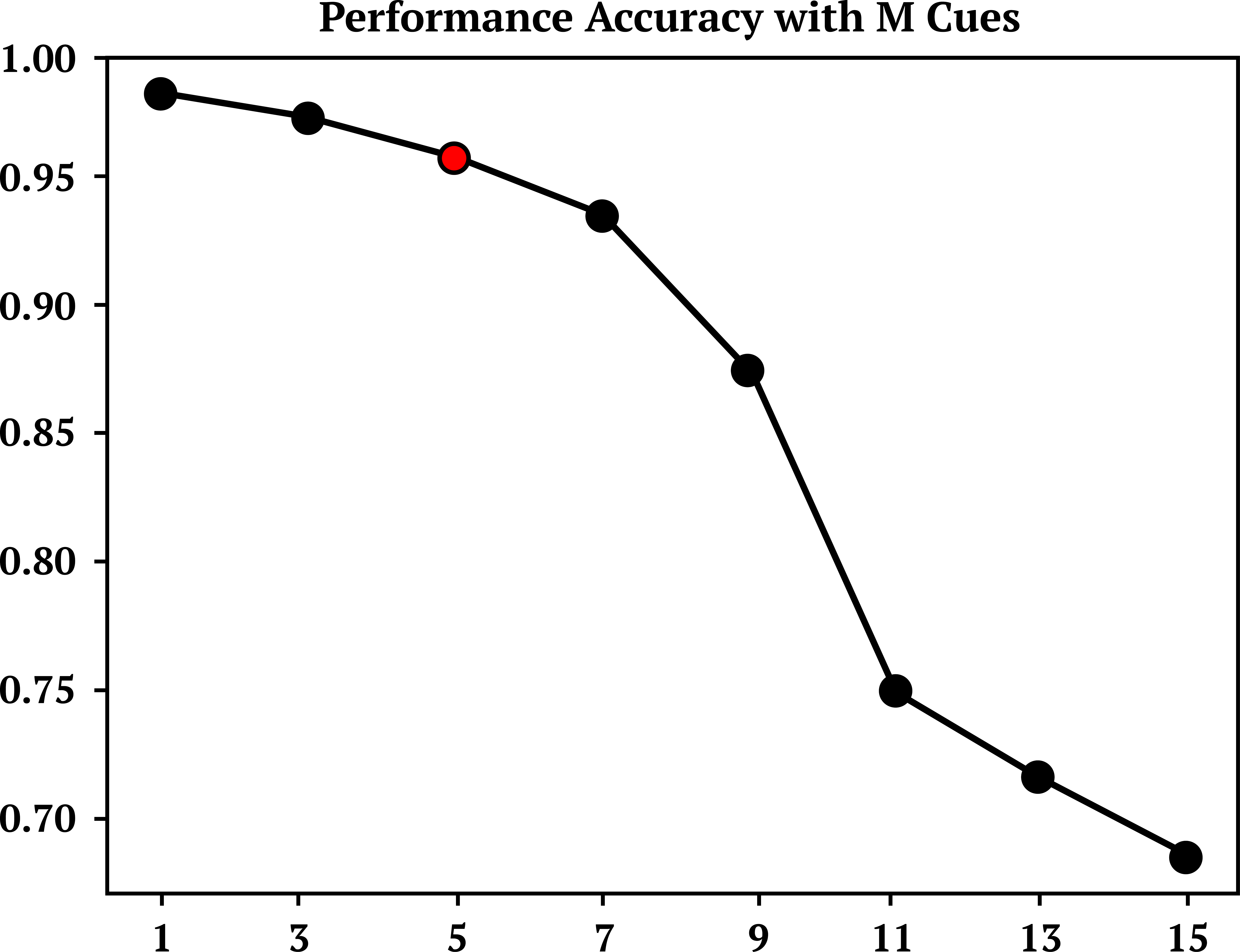}
    \end{center}
    \caption{\textbf{M Cue Performance.} Performance accuracy of cue association model trained on 5 cues and then tested on M cues between 1 and 15.}
    \label{figure:PerfGeneral}
\end{wrapfigure}

A representative trial of the one-shot cue association problem is shown in Figure \ref{figure:CueAssociation}. Performance on the testing set of novel cue permutations yields $95.6\%$ accuracy, which is averaged across 30 trainings with different randomly initialized parameters. Figure \ref{figure:PerfGeneral} demonstrates the performance accuracy of the network demonstrated in Figure \ref{figure:CueAssociation} when the number of cues presented, M, are varied from 1-15. Interestingly, while the network plasticity rule was only optimized for $M=5$ cues, the learning behavior exhibits the capability of accurately solving cue problems above and below the number of cues it was optimized for without additional training. Below $M=5$, $M=1$ obtains $98.1\%$ accuracy and $M=3$ obtains $96.7\%$ accuracy. Above $M=5$, there is a consistent loss in accuracy from $M=7$ with $94.2\%$ to $M=15$ with $68.7\%$. These results demonstrate that the learned neuromodulated plasticity rule generalizes in the task solving domain with respect to the number of cues without additional training on the meta-representation.








\begin{figure*}
    \centering
    \includegraphics[width=0.95\linewidth]{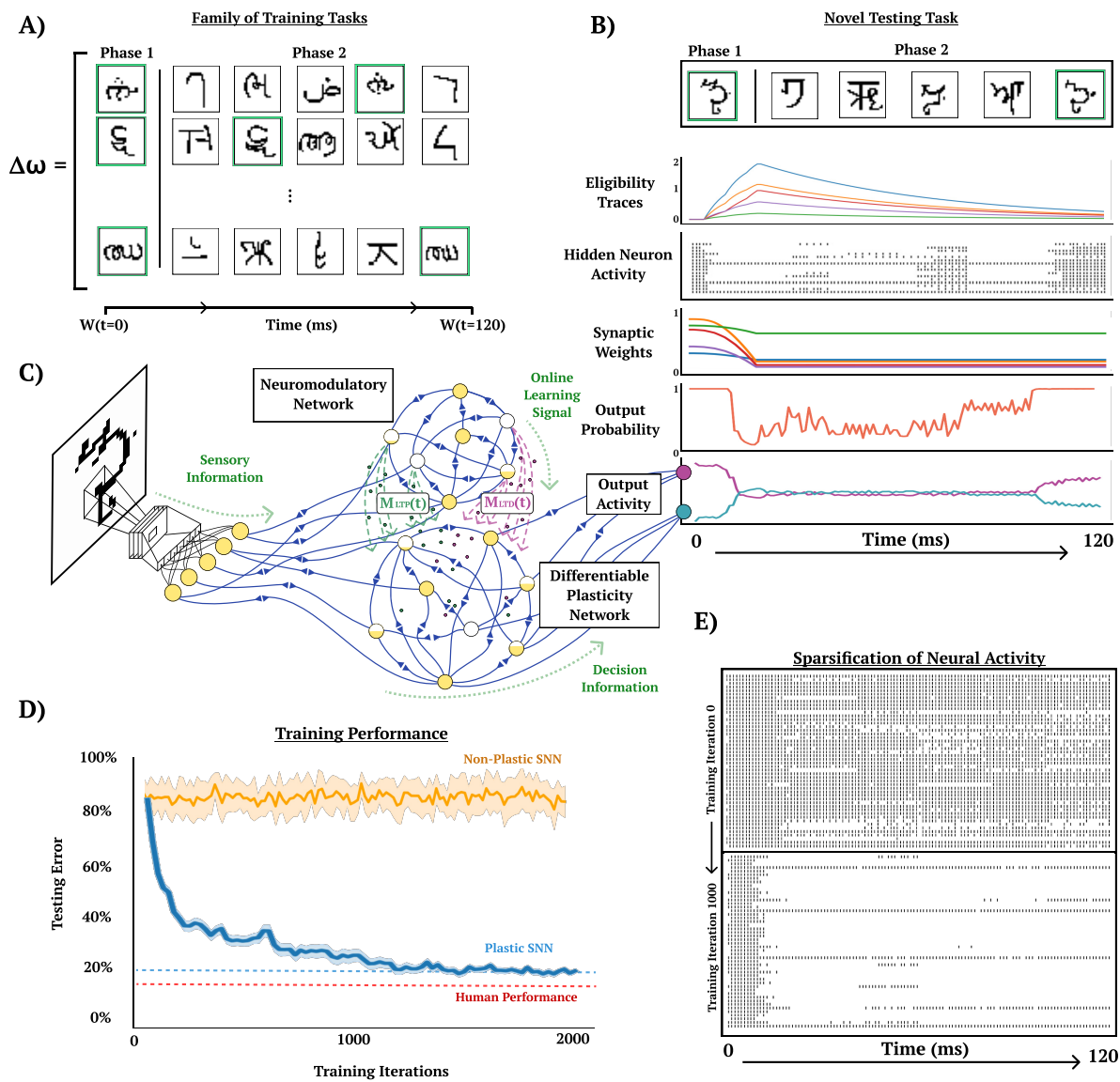}
    \caption{\textbf{One-shot character class recognition.} Visual demonstration of online one-shot character class recognition problem with triplet-based STDP. A) Examples of image sequences from three typical trials. Phase 1 image (green background) and samples of 5 phase 2 images, where character class from phase 1 image corresponds with a phase 2 image (green background) presented with random ordering. Meta-representation $\omega$ is updated based on the performance of $\omega$ on a family of training tasks. B) Activity of DP-SNN during the presentation of a novel testing sequence. C) Conceptual depiction of the interaction between CNN, NM-SNN and DP-SNN. Sensory information travels from CNN into both NM-SNN and DP-SNN, with DP-SNN receiving neuromodulatory signals from NM-SNN and outputting activity into classification neurons. D) Training performance comparison between plastic SNN (DP-SNN), non-plastic SNN, and human performer on testing set data. Human performer obtains ~15\%, DP-SNN 20.4\%, and non-plastic ~80\%. E) Depiction of the increased sparseness of hidden neuron activity as training progresses from training iteration 0 (top) to training iteration 1000 (bottom).}
    \label{figure:Omniglot}
\end{figure*}

\subsection*{Recognizing novel character classes from a single example}

Several learning challenges are presented in ref. \cite{lake2017building} with the aim of providing benchmarks that more closely demonstrate human-like intelligence in machines. The first among these challenges is the "characters challenge," which aims at benchmarking a learning algorithm's ability to recognize digits with few examples. The dataset for this challenge contains 1623 classes of handwritten characters across 50 unique alphabets, with each character consisting of 20 samples \cite{lake2015human}. In this challenge, a learner is presented with a phase 1 image as well as a set of phase 2 images (Figure \ref{figure:Omniglot}) where, one image presented is from the same phase 1 image class, and several other images presented are from other image classes. The phase 2 images are all presented simultaneously, and the learner must determine which image from phase 2 is in the phase 1 class. In the original design of this task, each image is able to be observed and compared simultaneously, and the image most closely matching the phase 1 image can be compared directly. A more challenging variation of this problem which aligns more closely to biological learning is presented in \cite{scherr2020one}, where each sample from phase 1 and phase 2 is presented sequentially instead of the learner being able to view and compare all samples simultaneously. The problem is considered solved correctly if the learner has the highest output activity for the image in phase 2 that matches the image class from phase 1. This variation of the characters challenge requires the learner to address the problem of holding information in memory across time and actively comparing that information with subsequently presented data, which even presents itself as a challenge for humans. Informal human testing from ref. \cite{scherr2020one} demonstrates error rates around 15\% based on 4 subjects and 100 trials.

\textbf{Experimental setup.} Both phase 1 and phase 2 images are presented for 20 ms with a simulation step size of 1 ms. One image is presented in phase 1 and five images are presented in phase 2 for a total trajectory time of 120 ms. This causes the character challenge to be particularly difficult because the set of testing tasks is much larger than the set of training tasks. It is argued that the character presentation should be intentionally small such that the learner must carry out spike-based computation and learning versus rate-based \cite{scherr2020one}. This time span is small compared to the average human visual reaction time which is around 331 ms \cite{jose2010comparison}. The phase 1 and phase 2 character classes are selected uniformly from a categorical distribution and the phase 2 characters are organized with random ordering. Neuromodulatory signals are only sent by the NM-SNN to the DP-SNN during the 20 ms presentation of the phase 1 character. During this period, the synapses must be modified to recognize the phase 2 image that belongs to the same character class as the phase 1 image. 

To increase the number of classes in the character dataset each character set is rotated by 90, 180, and 270 degrees, and are considered independent classes, increasing the number of character classes from 1623 to 6492. The character classes in the dataset are split into 20\% testing and 80\% training. There are $1.2\cdot10^{19}$ possible just on the ordering of character class arrangements in phase 2 in this problem, making it unlikely for the experiment to repeat any particular trial. Each gradient is computed across 256 cue trials and the model is updated for 2000 updates (Figure \ref{figure:Omniglot}D).

\begin{wrapfigure}[22]{L}{0.5\textwidth}
    \begin{center}
        \includegraphics[width=0.48\textwidth]{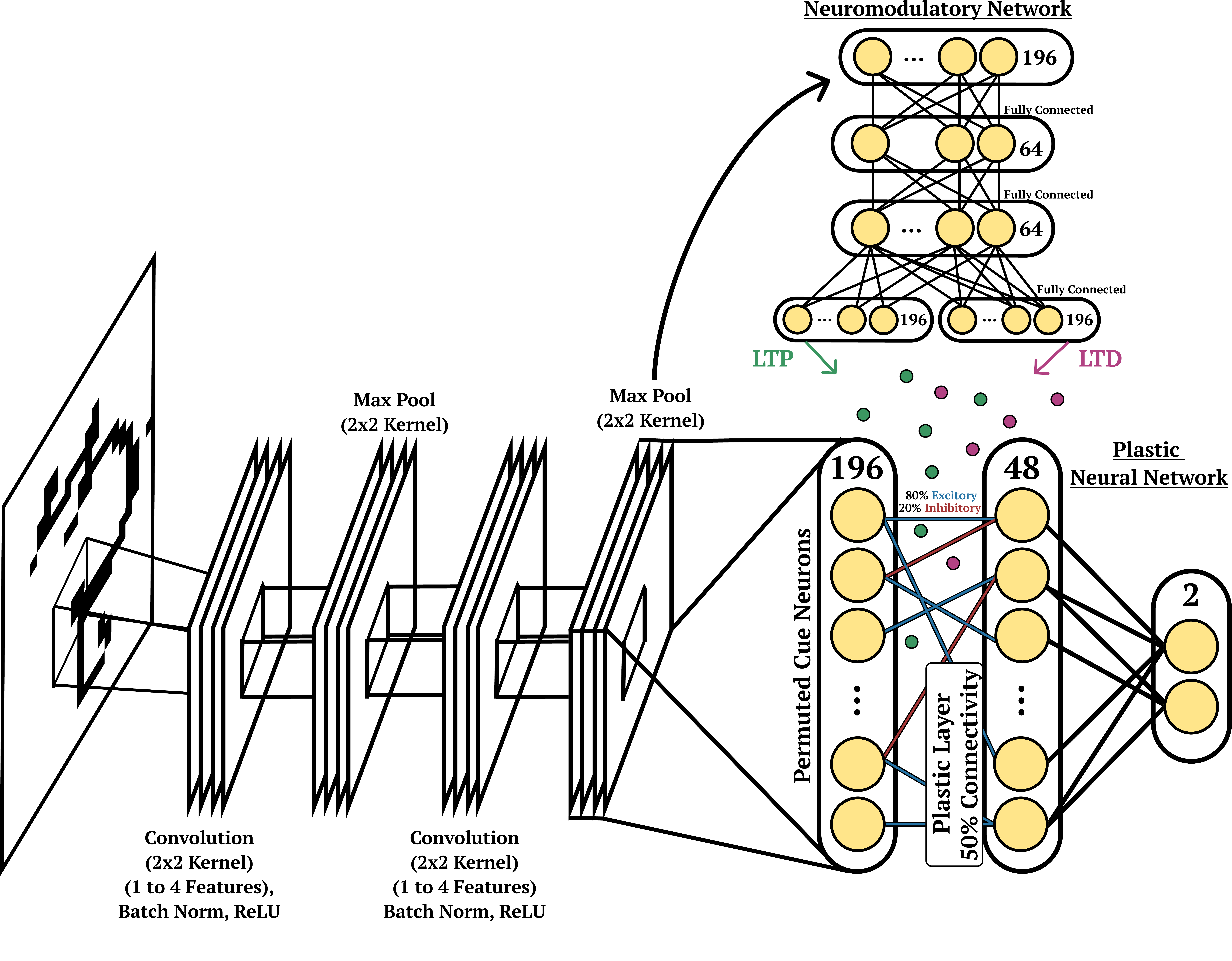}
    \end{center}
    \caption{\textbf{Character Recognition Architecture.} Depiction of the network structure for the DP-SNN (bottom, right), the NM-SNN (top), CNN pre-processing (bottom, left) for the cue association experiment.}
    \label{figure:NetDiagOmni}
\end{wrapfigure}

\textbf{Architecture.} The character image is fed into a several layers of a CNN for pre-processing and is flattened at the output. The flattened output is used as current input to a layer of 196 spiking neurons, which represent the input of the DP-SNN and the NM-SNN, see Figure \ref{figure:NetDiagOmni}. The DP-SNN consists of one hidden layer with 48 CUBA neurons (Methods, Equation \ref{eq:CurrentCUBA}-\ref{eq:VoltageCUBA}). Synaptic connections between the 196 input neurons and the 48 hidden layer neurons store LTP and LTD dynamics in separate eligibility traces based on an additive \textit{triplet} based STDP rule (Methods, Equation \ref{eq:TripletSTDP}). The triplet-based STDP provides a more accurate representation of biological STDP dynamics compared with the pair-based rule through the use of a \textit{slow} and \textit{fast} post-synaptic trace which accumulate post-synaptic activity with varied trace decay factors (Methods, Equation \ref{eq:TripletSTDPVariableX}). Connection probabilities between neurons are set to 50\% during initialization, with 20\% inhibitory synapses and 80\% excitatory. Modulatory signals are produced by the NM-SNN using an input neuron specific modulatory signal (Methods, Equation \ref{eq:ModEligibilityNeuronMod}) for both the LTP and LTD dynamics. The NM-SNN receives input from the image layer spiking neurons along with the DP-SNN hidden neuron activity. However, to make the challenge more difficult, the NM-SNN does not receive any additional inputs and must generate neuromodulation from the same sensory information as the DP-SNN. The NM-SNN consists of two layers of 64 CUBA neurons with fully-connected non-plastic synapses. The pre-processing CNN consists of the following steps: (1) convolution from 1 to 4 channels with a kernel size of 3, (2) batch norm, (3) ReLU operation, (4) max pooling with kernel and stride size of 2, (5) convolution from 4 to 4 channels with a kernel size of 3, (6) batch norm, (7) ReLU operation, and (8) a max pool with kernel and stride size of 2. This is then flattened and are used as current input to a 196 CUBA neurons which act as input for the DP-SNN and the NM-SNN.

\textbf{Results.} The performance of the NM-SNN and DP-SNN is compared to a non-plastic SNN using the same connective structure. The non-plastic SNN is demonstrated to be unable to solve this task with a testing error average of around $80\%$, which is equivalent to random selection. On the other hand, the DP-SNN obtains a testing error of $20.4\%$ after 2000 gradient steps on the outer-loop. This performance is comparable to informal human testing \cite{scherr2020one} which is around 15\%. A surprising finding was that the DP-SNN obtains $64.1\%$ accuracy on MNIST digits without any additional gradient steps on the plasticity parameters.

\section*{Discussion}


In this paper, we introduce a method for learning to learn with neuroscience models of synaptic plasticity in networks of spiking neurons, where the neuromodulated plasticity dynamics are learned through gradient descent and online learning tasks are solved with the learned neuromodulated plasticity dynamics online. This framework was demonstrated on two challenging online learning tasks: a one-shot continual learning problem and a one-shot image class recognition problem. These challenges required neuromodulated plasticity to act as the mechanism of intra-lifetime learning, and presented a way for learning the parameters of plasticity with gradient descent such that it can address these problems.

Previous work on the development of online SNN learning algorithms includes the work of e-prop \cite{bellec2020solution}, which is a plasticity rule that was mathematically derived from BPTT, where a learning signal defined by a given loss function over a task is projected to all neurons in the SNN using random feedback connections. This projected feedback interacts with an eligibility trace that accumulates the BPTT plasticity approximation to update synaptic weights. E-prop was demonstrated to be competitive with BPTT on several temporal learning benchmarks. In ref. \cite{scherr2020one}, a method called \textit{natural e-prop} is introduced, which uses the plasticity dynamics of e-prop and learns a neuromodulatory signal toward solving several one-shot learning challenges. Another online learning algorithms for SNNs is Surrogate-gradient Online Error triggered Learning (SOEL) \cite{stewart2020online}. SOEL calculates a global error signal and uses surrogate gradient descent to create a plasticity-like rule for updating the network synapses online. Works like e-prop and SOEL are not competing algorithms, but rather are complimentary with respect to this framework. The e.g. timing parameters, voltage parameters, and, the surrogate gradient parameters could be learned by gradient descent using these methods as the inner-loop optimization to produce an even more effective version of the existing algorithm. There have also been many previous contributions toward neuromodulated plasticity in non-spiking Artificial Neural Networks (ANNs) \cite{soltoggio2008evolutionary, miconi2020backpropamine, risi2012unified, velez2017diffusion, beaulieu2020learning}. However, plastic ANNs have been demonstrated to struggle maintaining functional stability across time due to their continuous nature which causes synapses to be in a constant state of change \cite{schmidgall2021stable}. The effect of this instability was shown to not disturb the performance as significantly in plastic SNNs as it did on plastic ANNs.

To realize the full potential of this framework, described here are several topics for future research, including: incorporating cell-type specific neuromodulatory signals \cite{doi:10.1073/pnas.2111821118} into the learning process; exploring the addition of glial cell dynamics \cite{ivanov2021increasing, gordleeva2021modelling}; providing deeper insight into the learning capabilities of different plasticity rules in the neuroscience literature, such as the wide-range of existing voltage-dependent plasticity rules, rate-based plasticity rules, and spike-timing dependent plasticity rules; and exploring the use of this framework on robotic and reinforcement learning experiments. Another direction might explore learning the neural architecture in conjunction with the plasticity parameters, since architecture is known to play a significant role in the function of neural dynamics \cite{gaier2019weight}. Recent works have explored learning the plasticity rule equation in addition to the plasticity rule parameters \cite{jordan2021evolving}. A differentiable plasticity rule search constrained toward biological realism may provide more powerful learning applications of this framework.

Finally, addressing the problem of online learning has been a central focus of neuromorphic computing \cite{davies2021advancing}. The existing need for learning methods that can be used on these systems has impeded the use of neuromorphic systems in real-world applications. From a practical perspective, backpropagation on these systems is only envisioned as a utility for offline training since on-chip BPTT is expensive with respect to complexity, memory, and energy efficiency, and is not naturally suited for online learning. Instead, some neuromorphic systems have invested in \textit{on-chip plasticity} in part to address online learning in hopes that an effective method for utilizing this capability is discovered. Neuromorphic processors implement \textit{on-chip plasticity} by allowing the flexible reconfiguration of a set of local variables that interact to adapt synaptic weights \cite{davies2021advancing, davies2018loihi, jin2010implementing, rajendran2019low, pehle2022brainscales}. The reconfiguration of these variables have historically modelled learning rules from the neuroscience literature. In spite of this, the goal of finding learning rules that can solve a wide variety of challenging problems (like backpropagation) while building off of the impressive capabilities of the brain remains open. We hope that this framework of learning to learn with backpropagation inspires the next generation of on-chip learning algorithms for the field of neuromorphic computing.

The framework of learning to learn with neuromodulated synaptic plasticity in this paper provides a method for combining the power of gradient descent with neuroscience models of plasticity, which opens the doors toward a better synthesis of machine learning and neuroscience.

\section*{Methods}


\subsection*{Leaky Integrate-and-Fire}

The Leaky Integrate-and-Fire (LIF) neuron model is a phenomenological model of neural firing-dynamics. Activity is integrated into the neuron and stored across time, and, once the stored activity surpasses a threshold value, a binary signal is emitted and the voltage is reset. The "leaky" part of the model name refers to an introduced time-dependent decay dynamic acting on the membrane potential. While the simplicity of the LIF dynamics deviates from the complexity of the biological neuron, the purpose of the model is to capture the essence of neuron dynamics while providing value from a computational perspective. The LIF neuron model requires among the fewest computational operations to implement compared with other neuron models. 

To begin describing the LIF dynamics, we represent the continuous difference equation $\tau\frac{d\textbf{\textit{v}}_{j}}{dt}$ for the voltage state $\textbf{\textit{v}}_{j}(t) \in \mathbb{R}$ as a discrete time equation $\textbf{\textit{v}}_{j} + \tau\frac{d\textbf{\textit{v}}_{j}}{dt} = \textbf{\textit{v}}_{j}(t+\Delta\tau)$ since computational models of spiking neurons typically operate across discrete update intervals.

\begin{equation}\label{eq:LIF}
    \textbf{\textit{v}}_{j}(t+\Delta\tau) = \textbf{\textit{v}}_{j}(t) - \alpha_{\textbf{\textit{v}}} [\textbf{\textit{v}}_{j}(t) - v_{rest}] + R\textbf{\textit{I}}_{j}(t),
\vspace{1.3mm}
\end{equation}

The term $\alpha_{\textbf{\textit{v}}} [\textbf{\textit{v}}_{j}(t) - v_{rest}]$ represents the membrane potential leak, where $\alpha_{\textbf{\textit{v}}} \in [0, 1]$ is the leak time-constant and $v_{rest} \in \mathbb{R}$ as the neuron resting potential, which is the value that the membrane potential returns to in the absence of external activity. $\textbf{\textit{I}}_{j}(t) \in \mathbb{R}$ represents incoming current, which is the source of an increase in voltage $\textbf{\textit{v}}_{j}(t)$ for the neuron $j$. This current is scaled by a resistance factor $R \in \mathbb{R}$.

\begin{equation}\label{eq:Spike}
    \textbf{\textit{s}}_{j}(t) = H(\textbf{\textit{v}}_{j}(t)) = \begin{dcases}
        0 & \textbf{\textit{v}}_{j}(t) \leq v_{th} \\
        1 & \textbf{\textit{v}}_{j}(t) > v_{th} \\
    \end{dcases}
,
\vspace{1.3mm}
\end{equation}

$H: \mathbb{R} \to \{0, 1\}$ is a piece-wise step function which, in the case of a spiking neuron, outputs $1$ when a neuron's membrane potential surpasses the defined firing threshold $v_{th} \in \mathbb{R}$ and otherwise outputs 0. In the LIF neuron model, once a neuron fires a spike, the membrane potential is reset to its resting potential $\textbf{\textit{v}}_{j}(t) \leftarrow v_{rest}$.

In a spiking neural network, $\textbf{\textit{I}}_{j}(t)$ from Equation \ref{eq:LIF} is defined as $\textbf{\textit{I}}_{j} = \sum_{i}\textbf{\textit{W}}_{i,j}\textbf{\textit{s}}_{i}(t)$, which represents the sum of weighted spikes from all pre-synaptic neurons $i$ that are connected to post-synaptic neuron $j$. The weight of each spike is given by $\textbf{\textit{W}}_{i,j}(t) \in \mathbb{R}$, with $\textbf{\textit{W}}_{i,j}(t) < 0$ representing inhibitory connections, and $\textbf{\textit{W}}_{i,j}(t) > 0$ representing excitatory connections. 

\begin{equation}\label{eq:ReducedLIF}
    \textbf{\textit{v}}_{j}(t+\Delta\tau) = \textbf{\textit{v}}_{j}(t) - \alpha_{v} [\textbf{\textit{v}}_{j}(t) - v_{rest}] + R\sum_{i}\textbf{\textit{W}}_{i,j}(t)\textbf{\textit{s}}_{i}(t),
\vspace{1.3mm}
\end{equation}

Consistent with Intel's neuromorphic processor code named Loihi, our experiments use an adaptation of the LIF which incorpoates current called the Current-based Leaky-integrate and fire (CUBA) neuron model \cite{davies2018loihi}.

\begin{equation}\label{eq:CurrentCUBA}
    \textbf{\textit{u}}_{i}(t+\Delta\tau) = \textbf{\textit{u}}_{i}(t) - \alpha_{u} [\textbf{\textit{u}}_{i}(t) - u_{rest}] + \sum_{j}\textbf{\textit{W}}_{i,j}(t)\textbf{\textit{s}}_{j}(t),
\vspace{1.3mm}
\end{equation}

\begin{equation}\label{eq:VoltageCUBA}
    \textbf{\textit{v}}_{i}(t+\Delta\tau) = \textbf{\textit{v}}_{i}(t) - \alpha_{v} [\textbf{\textit{v}}_{i}(t) - v_{rest}] + R\textbf{\textit{u}}_{i}(t).
\vspace{1.3mm}
\end{equation}

In the CUBA neuron model, a decaying current trace $\textbf{\textit{u}}_{i}(t)$ integrates incoming current $\textbf{\textit{I}}_{j} = \sum_{i}\textbf{\textit{W}}_{i,j}\textbf{\textit{s}}_{i}(t)$ from pre-synaptic neurons $i$ into the post-synaptic current trace $j$  in Equation \ref{eq:CurrentCUBA}. Then instead of current $\textbf{\textit{I}}_{j}$ directly modifying the neuron membrane potential $\textbf{\textit{v}}_{i}(t)$  in Equation \ref{eq:VoltageCUBA}, the current trace $\textbf{\textit{u}}_{i}(t)$ takes its place.

\textbf{Backpropagation through spiking neurons.} The role of $H(\cdot)$ in Equation \ref{eq:Spike} can be viewed analogously to the non-linear activation function used in artificial neural networks. However, unlike most utilized non-linearities, $H(\cdot)$ is non-differentiable, and hence backpropagating gradients becomes particularly challenging. To backpropagate through the non-differentiable function $H(\cdot)$, Spike Layer Error Reassignment in Time (SLAYER) is used. SLAYER represents the derivative of the spike function $H(\cdot)$ with a surrogate gradient, and backpropagates error through a temporal credit assignment policy \cite{shrestha2018slayer}.

\subsection*{Spike-timing based Plasticity Rules}

Spike-timing Dependent Plasticity rules, unlike rate-based models, are dependent on the relationship between precise spike-timing events in pre- and post-synaptic neurons \cite{gerstner1996neuronal, markram1997regulation, bi1998synaptic, sjostrom2001rate}. Equations for neuronal and synaptic plasticity dynamics are presented as discrete-time update equations as opposed to continuous-time equations to provide a closer correspondence to the computational implementation.




\subsubsection*{Synaptic Traces}



STDP can be defined as an iterative update rule through the use of synaptic activity traces.

\begin{equation}\label{eq:STDPVariableX}
\textbf{\textit{x}}_{i}^{(l)}(t+\Delta\tau) = \alpha_{x} \textbf{\textit{x}}_{i}^{(l)}(t) + f(\textbf{\textit{x}}_{i}^{(l)}(t)) \textbf{\textit{s}}^{(l)}_{i}(t).
\vspace{1.3mm}
\end{equation}

The bio-physical meaning of the activity trace $\textbf{\textit{x}}_{i}^{(l)}(t) \in \mathbb{R} > 0$ is left abstract, as there are several candidates for the representation of this activity. For pre-synaptic events, this quantity could represent the amount of bound glutamate or the quantity of activated NMDA receptors, and for post-synaptic events the synaptic voltage by a backpropagating action potential or by calcium entry through a backpropagating action potential. 

The variable $\alpha_{x} \in (0, 1)$ is traditionally represented as a quantity $(1-\frac{1}{\tau}$), which decays the activity trace to zero at a rate inversely proportional to the magnitude of the time constant $\tau \in \mathbb{R} > 1$. The trace $\textbf{\textit{x}}_{i}^{(l)}(t)$ is updated by a quantity proportional to $f: \mathbb{R} \to \mathbb{R}$ in the presence of a spike $\textbf{\textit{s}}^{(l)}_{i}(t)$. This synaptic trace is referred to as an all-to-all synaptic trace scheme since each pre-synaptic spike is paired with every post-synaptic spike in time indirectly via the decaying trace.

In the linear case of this update rule $f(\textbf{\textit{x}}_{i}^{(l)}(t)) = \boldsymbol\beta \in \mathbb{R} > 0$, the trace is updated by a constant factor $\boldsymbol\beta$ in the presence of a spike $\textbf{\textit{s}}^{(l)}_{i}(t)$.

\begin{equation}\label{eq:LinearSTDPVariableX}
\textbf{\textit{x}}_{i}^{(l)}(t+\Delta\tau) = \alpha_{x} \textbf{\textit{x}}_{i}^{(l)}(t) + \boldsymbol\beta \textbf{\textit{s}}^{(l)}_{i}(t).
\vspace{1.3mm}
\end{equation}

Another candidate for the function $f(\textbf{\textit{x}}_{i}^{(l)}(t))$ is $\boldsymbol\beta[x_{max} - \textbf{\textit{x}}_{i}^{(l)}(t)]$, which updates the trace by a constant $\boldsymbol\beta$ together with a factor $[x_{max} - \textbf{\textit{x}}_{i}^{(l)}(t)]$ that scales the update depending on the relationship between $\textbf{\textit{x}}_{i}^{(l)}(t)$ and its proximity to the trace saturation point $x_{max} \in \mathbb{R} > 0$ \cite{morrison2008phenomenological}.

\begin{equation}\label{eq:BioSTDPVariableX}
\textbf{\textit{x}}_{i}^{(l)}(t+\Delta\tau) = \alpha_{x} \textbf{\textit{x}}_{i}^{(l)}(t) + \boldsymbol\beta [x_{max} - \textbf{\textit{x}}_{i}^{(l)}(t)]  \textbf{\textit{s}}^{(l)}_{i}(t).
\vspace{1.3mm}
\end{equation}

When $\boldsymbol\beta < 1$ , as $\textbf{\textit{x}}_{i}^{(l)}(t)$ approaches $x_{max}$, the update scale $[x_{max} - \textbf{\textit{x}}_{i}^{(l)}(t)]$ reduces the magnitude of the trace update, producing a soft-bounded range $0 \leq \textbf{\textit{x}}_{i}^{(l)}(t) \leq x_{max}$.

\subsubsection*{Pair-based STDP}



The pair-based model of STDP describes a plasticity rule from which synapses are changed as a product of the timing relationship between \textit{pairs} of pre- and post-synaptic activity.


\begin{equation}\label{eq:PairbasedSTDP}
\textbf{\textit{W}}_{i,j}^{(l)}(t+\Delta\tau) = \textbf{\textit{W}}_{i,j}^{(l)}(t) +\\
 \textbf{\textit{A}}_{+,i,j}(\textbf{\textit{W}}_{i,j}^{(l)}(t)) \textbf{\textit{x}}_{i}^{(l-1)}(t) \textbf{\textit{s}}_{j}^{(l)}(t) -\\
 \textbf{\textit{A}}_{-,i,j}(\textbf{\textit{W}}_{i,j}^{(l)}(t)) \textbf{\textit{x}}_{j}^{(l)}(t) \textbf{\textit{s}}_{i}^{(l-1)}(t).
\end{equation}

Weight potentiation is realized in the presence of a post-synaptic firing $\textbf{\textit{s}}_{j}^{(l)}(t) = 1$ by a quantity proportional to the pre-synaptic trace $\textbf{\textit{x}}_{i}^{(l-1)}(t)$. Likewise, weight depression is realized in the presence of a pre-synaptic $\textbf{\textit{s}}_{i}^{(l-1)}(t) = 1$ proportional to the post-synaptic trace  $\textbf{\textit{x}}_{j}^{(l)}(t)$.  Potentiation and depression are respectively scaled by $\textbf{\textit{A}}_{+,i,j}: \mathbb{R} \to \mathbb{R}$ and $\textbf{\textit{A}}_{-,i,j}: \mathbb{R} \to \mathbb{R}$, which are functions that characterize the update dependence on the current weight of the synapse $\textbf{\textit{W}}_{i,j}^{(l)}(t)$. Hebbian pair-based STDP models generally define $\textbf{\textit{A}}_{+,i,j}(\textbf{\textit{W}}_{i,j}^{(l)}(t)) > 0$ and $\textbf{\textit{A}}_{-,i,j}(\textbf{\textit{W}}_{i,j}^{(l)}(t)) > 0$, whereas anti-Hebbian models define $\textbf{\textit{A}}_{+,i,j}(\textbf{\textit{W}}_{i,j}^{(l)}(t)) < 0$ and $\textbf{\textit{A}}_{-,i,j}(\textbf{\textit{W}}_{i,j}^{(l)}(t)) < 0$. 

\subsubsection*{Weight-dependence}

An additive model of pair-based STDP defines $\textbf{\textit{A}}_{+,i,j}(\textbf{\textit{W}}_{i,j}^{(l)}(t)) = \boldsymbol\eta_{+,i,j}^{(l)}$, which scales LTP and LTD linearly by a factor $\boldsymbol\eta_{+,i,j}^{(l)} \in \mathbb{R}$ and $\boldsymbol\eta_{-,i,j}^{(l)} \in \mathbb{R}$ respectively.




\begin{equation}\label{eq:STDPAdd}
\textbf{\textit{W}}_{i,j}^{(l)}(t+\Delta\tau) = \textbf{\textit{W}}_{i,j}^{(l)}(t) + \boldsymbol\eta_{+,i,j}^{(l)} \textbf{\textit{x}}_{i}^{(l-1)}(t) \textbf{\textit{s}}_{j}^{(l)}(t) - \boldsymbol\eta_{-,i,j}^{(l)} \textbf{\textit{x}}_{j}^{(l)}(t) \textbf{\textit{s}}_{i}^{(l-1)}(t).
\end{equation}

Additive models of STDP demonstrate strong synaptic competition, and hence tend to produce clear synaptic specialization \cite{10.1371/journal.pone.0025339}. However, without any dependence on the weight parameter for regulation, the weight dynamics may grow either without bound or, with hard bounds, bimodally \cite{morrison2008phenomenological, 10.1371/journal.pone.0025339, rubin_stdpstability}.



A multiplicative, or weight dependent, model of pair-based STDP defines  $\textbf{\textit{A}}_{+,i,j}(\textbf{\textit{W}}_{i,j}^{(l)}(t)) = \boldsymbol\eta_{+,i,j}(\textbf{\textit{W}}_{max} - \textbf{\textit{W}}_{i,j}^{(l)}(t))$ for LTP, which scales the effect of potentiation based on the proximity of the weight $\textbf{\textit{W}}_{i,j}^{(l)(t)}$ to the defined weight soft upper-bound $\textbf{\textit{W}}_{max}$. Similarly, LTD defines $\textbf{\textit{A}}_{-,i,j}(\textbf{\textit{W}}_{i,j}^{(l)}(t)) = \boldsymbol\eta_{-,i,j}(\textbf{\textit{W}}_{i,j}^{(l)}(t) - \textbf{\textit{W}}_{min})$, which scales weight depression according to the defined soft-lower bound $\textbf{\textit{W}}_{min}$.


\begin{equation}\label{eq:STDPMult}
\textbf{\textit{W}}_{i,j}^{(l)}(t+\Delta\tau) = \textbf{\textit{W}}_{i,j}^{(l)}(t) + \boldsymbol\eta_{+,i,j}(\textbf{\textit{W}}_{max} - \textbf{\textit{W}}_{i,j}^{(l)}(t)) \textbf{\textit{x}}_{i}^{(l-1)}(t) \textbf{\textit{s}}_{j}^{(l)}(t) - \boldsymbol\eta_{-,i,j}(\textbf{\textit{W}}_{i,j}^{(l)}(t) - \textbf{\textit{W}}_{min}) \textbf{\textit{x}}_{j}^{(l)}(t) \textbf{\textit{s}}_{i}^{(l-1)}(t).
\end{equation}

LTP and LTD produce weight changes depending on their relationship to the upper- and lower-bound, with LTP more effective when weights are farther from the upper-bound and LTD more effective when weights are farther from the lower bound. The use of soft bounds in practice leads to LTD dominating over LTP \cite{10.1371/journal.pone.0025339, rubin_stdpstability,stablehebbian, PhysRevE.59.4498,song2000competitive} and, opposite to additive pair-based STDP, fails to demonstrate clear synaptic specialization \cite{10.1371/journal.pone.0025339}.

Additive and multiplicative models of STDP have been regarded as extremes among a range of representations, with LTP as $\textbf{\textit{A}}_{+,i,j}(\textbf{\textit{W}}_{i,j}^{(l)}(t)) = \boldsymbol\eta_{+,i,j}(\textbf{\textit{W}}_{max} - \textbf{\textit{W}}_{i,j}^{(l)}(t))^{\mu}$ and LTD as $\textbf{\textit{A}}_{-,i,j}(\textbf{\textit{W}}_{i,j}^{(l)}(t)) = \boldsymbol\eta_{-,i,j}(\textbf{\textit{W}}_{i,j}^{(l)}(t) - \textbf{\textit{W}}_{min})^{\mu}$ \cite{10.1371/journal.pbio.0030068, gutig2003learning}. Here, the parameter $\mu$ acts as an exponential weight-dependence scale, with $\mu = 0$ producing an additive model, and $\mu = 1$ producing a multiplicative model. Values of $0 < \mu < 1$ result in rules with intermediate dependence on $\textbf{\textit{W}}_{i,j}^{(l)}(t)$.

\subsubsection*{Triplet-based STDP}



Experimental data has demonstrated that pair-based STDP models cannot provide an accurate representation of biological STDP dynamics under certain conditions. Particularly, these rules cannot reproduce triplet and quadruplet experiments, and also cannot account for the frequency-dependence of plasticity demonstrated in STDP experiments \cite{sjostrom2001rate, senn2001algorithm}.

\begin{equation}\label{eq:TripletSTDPVariableX}
\textbf{\textit{x}}_{i, \tau}^{(l)}(t+\Delta\tau) = \alpha_{\tau} \textbf{\textit{x}}_{i, \tau}^{(l)}(t) + f(\textbf{\textit{x}}_{i, \tau}^{(l)}(t))\textbf{\textit{s}}^{(l)}_{i}(t)
\vspace{1.3mm}
\end{equation}

To address the representation limitations of pair-based STDP, a plasticity rule based on a triplet interaction between one pre-synaptic spike and two post-synaptic spikes is proposed in ref. \cite{gjorgjieva2011triplet}. To implement this, a second \textit{slow} synaptic trace is introduced for the post-synaptic neuron is introduced, with a time constant $\alpha_{\tau} \in \mathbb{R} > \alpha_{x}$, with $\alpha_{x}$ representing the decay rate of the \textit{fast} synaptic trace from Equation \eqref{eq:STDPVariableX}. More specifically, the triplet model of STDP produces \textit{LTP} dynamics that are dependent on the pre-synaptic trace  $\textbf{\textit{x}}_{i}^{(l-1)}(t)$ (Equation \eqref{eq:TripletSTDPVariableX}) and the \textit{slow} post-synaptic trace $\textbf{\textit{x}}_{j, \tau}^{(l)}(t - \Delta\tau)$, which is evaluated at time $t - \Delta\tau$, one timestep prior to the evaluation of traces $\textbf{\textit{x}}_{j}^{(l)}(t)$ and $\textbf{\textit{x}}_{i}^{(l-1)}(t)$:

\begin{equation}\label{eq:TripletSTDP}
\textbf{\textit{W}}_{i,j}^{(l)}(t+\Delta\tau) = \textbf{\textit{W}}_{i,j}^{(l)}(t) + \textbf{\textit{A}}_{+,i,j}(\textbf{\textit{W}}_{i,j}^{(l)}(t)) \textbf{\textit{x}}_{i}^{(l-1)}(t) \textbf{\textit{x}}_{j, \tau}^{(l)}(t - \Delta\tau) - \textbf{\textit{A}}_{-,i,j}(\textbf{\textit{W}}_{i,j}^{(l)}(t)) \textbf{\textit{x}}_{j}^{(l)}(t) \textbf{\textit{s}}_{i}^{(l)}(t)
\end{equation}

The triplet rule has demonstrated to explain several plasticity experiments more effectively than pair-based STDP \cite{sjostrom2001rate, wang2005coactivation, gjorgjieva2011triplet}. Additionally, the triplet rule has been demonstrated to be capable of being mapped to the BCM rule under the assumption that (1) pre- and post-synaptic spiking behavior assumes independent stochastic spike trains, (2) LTD is produced in the presence of low post-synaptic firing rates, (3) LTP is produced in the presence of high post-synaptic firing rates, and (4) the triplet term is dependent on the average post-synaptic firing frequency \cite{pfister2006triplets}. If these requirements are matched, the presented triplet-based STDP rule demonstrates the properties of the BCM rule, such as synaptic competition which produces input selectivity, a requirement for receptive field development \cite{bienenstock1982theory, pfister2006triplets}.

\subsection*{Neuromodulatory Plasticity Rules}

Synaptic learning rules in the context of SNNs mathematically describe the change in synaptic strength between a pre-synaptic neuron \textit{i} and post-synaptic neuron $j$. At the biological level, these changes are products of complex dynamics between a diversity of molecules interacting at multiple time-scales. Many behaviors require the interplay of activity on the time-scale of seconds to minutes, such as exploring a maze, and on the time-scale of milliseconds, such as neuronal spiking. Learning rules must be capable of effectively integrating these two diverse time-scales. Thus far, the learning rules observed have been simplified to equations which modify the synaptic strength $\textbf{\textit{W}}_{i,j}^{(l)}(t)$ based on local synaptic activity without any motivating guidance and without the presence of external modulating factors. 

Biological experiments have demonstrated that synaptic plasticity is often dependent on the presence of neuromodulators such as dopamine \cite{steinberg2013causal, schultz1993responses, seamans2007dopamine, zhang2009gain, Speranza2021DopamineTN}, noradrenaline \cite{ranganath2003neural, salgado2012noradrenergic}, and acetylcholine \cite{ranganath2003neural, teles2013cholinergic, brzosko2015retroactive, hasselmo2006role, zannone2018acetylcholine, Zannone2018AcetylcholinemodulatedPI, Hasselmo2006TheRO}. These modulators often act to regulate plasticity at the synapse by gating synaptic change, with recent evidence suggesting that interactions more complex than gating occur \cite{zhang2009gain, fremaux2016neuromodulated, gerstner2018eligibility}. The interaction between neuromodulators and eligibility traces has served as an effective paradigm for many biologically-inspired learning algorithms \cite{fremaux2013reinforcement, bellec2020solution, bellec2019biologically, bing2018end, seung2003learning}.


\subsubsection*{Eligibility Traces}

Rather than directly modifying the synaptic weight, local synaptic activity leaves an activity flag, or eligibility trace, at the synapse. The eligibility trace does not immediately produce a change, rather, weight change is realized in the presence of an additional signal. In the theoretical literature on three-factor learning, this signal has been theorized to be accounted for by \textit{external}, or non-local, activity \cite{MARDER20121, fremaux2016neuromodulated, gerstner2018eligibility}. For learning applications, this third signal could be a prediction error, or for reinforcement learning, an advantage prediction \cite{fremaux2013reinforcement}. In a Hebbian learning rule, the eligibility trace can be described by the following equation:

\begin{equation}\label{eq:Eligibility}
\textbf{\textit{E}}_{i,j}^{(l)}(t+\Delta\tau) = \gamma \textbf{\textit{E}}_{i,j}^{(l)}(t) + \boldsymbol\alpha_{i,j}f_{i}(x^{(l-1)}_{i})g_{j}(x^{(l)}_{j}).
\end{equation}

The constant $\gamma \in [0, 1]$ inversely determines the rate of decay for the trace, $\boldsymbol\alpha_{i,j} \in \mathbb{R}$ is a constant determining the rate at which activity trace information is introduced into the eligibility trace, $f_{i}$ is a function of pre-synaptic activity $x^{(l-1)}_{i}$, and $g_{j}$ a function of post-synaptic activity $x^{(l)}_{j}$. These functions are indexed by their corresponding pre- and post-synaptic neuron $i$ and $j$ since the synaptic activity eligibility dynamics may be dependent on neuron type or the region of the network.

Both rate- and spike-based models of plasticity can be represented with the eligibility trace dynamics described in Equation \eqref{eq:Eligibility}. Spike-based models of plasticity, such as the triplet-based (Equation \ref{eq:TripletSTDP}) and pair-based model (Equation \ref{eq:PairbasedSTDP}), often require two synaptic flags $\textbf{\textit{E}}_{+,i,j}^{(l)}$ and $\textbf{\textit{E}}_{-,i,j}^{(l)}$ for LTP and LTD respectively.


\subsubsection*{Modulatory Eligibility Traces}

In the theoretical literature, eligibility traces alone are not sufficient to produce a change in synaptic efficacy \cite{gerstner2018eligibility, fremaux2016neuromodulated}. Instead, weight changes are realized in the presence of a third signal.

\begin{equation}\label{eq:ModEligibility}
\textbf{\textit{W}}_{i,j}^{(l)}(t+\Delta\tau) = \textbf{\textit{W}}_{i,j}^{(l)}(t) + M(t) \textbf{\textit{E}}_{i,j}^{(l)}(t).
\end{equation}

Here, $M(t) \in \mathbb{R}$ acts as a global third signal which is referred to as a neuromodulator. Weight changes no longer occur in the absence of the neuromodulatory signal, $M(t) = 0$. When the value $M(t)$ ranges from positive to negative values, the \textit{magnitude} and \textit{direction} of change is determined causing LTP and LTD to both scale and reverse in the presence of certain stimuli. 

The interaction between individual neurons and the global neuromodulatory signal need not be entirely defined multiplicatively as in Equation \ref{eq:ModEligibility}, but can have neuron-specific responses defined by the following dynamics:

\begin{equation}\label{eq:ModEligibilityNeuron}
\textbf{\textit{W}}_{i,j}^{(l)}(t+\Delta\tau) = \textbf{\textit{W}}_{i,j}^{(l)}(t) + h_{j}(M(t)) \textbf{\textit{E}}_{i,j}^{(l)}(t).
\end{equation}

The function $h_{j}: \mathbb{R} \to \mathbb{R}$ is a neuron-specific response function which determines how the post-synaptic neuron $j$ responds to the neuromodulatory signal $M(t)$. This form of neuromodulation accounts for random-feedback networks when $h_{j}(M(t)) = h(b_{j}M(t))$. However, this form of neuromodulation does not account for the general supervised learning paradigm through backpropagating error. Equation \eqref{eq:ModEligibilityNeuron} must be extended to account for neuron-specific neuromodulatory signals:

\begin{equation}\label{eq:ModEligibilityNeuronMod}
\textbf{\textit{W}}_{i,j}^{(l)}(t+\Delta\tau) = \textbf{\textit{W}}_{i,j}^{(l)}(t) + M_{j}(t) \textbf{\textit{E}}_{i,j}^{(l)}(t).
\end{equation}

In \textit{layered} networks being optimized through backpropagation, the neuron-specific error is $M_{j}(t)$. In the case of backpropagation, $M_{j}(t)$ is calculated as a weighted sum from the errors in the neighboring layer closest to the output. The neuron-specific error in Equation \ref{eq:ModEligibilityNeuronMod} can also be computed with the dimensionality of the pre-synaptic neurons, $M_{i}(t)$, which was the form of neuromodulation used in both experiments from the Experiments section.

\section*{Declaration of Competing Interest}

The authors declare that they have no known competing financial interests or personal relationships that could have appeared to influence the work reported in this paper.

\section*{Acknowledgments}

The program is funded by Office of the Under Secretary of Defense (OUSD) through the Applied Research for Advancement of S\&T Priorities (ARAP) Program work unit 1U64.



 
\bibliographystyle{unsrt}
 \bibliography{cas-refs}

\begin{thebibliography}{10}

\bibitem{doi:10.1073/pnas.1611835114}
James Kirkpatrick, Razvan Pascanu, Neil Rabinowitz, Joel Veness, Guillaume
  Desjardins, Andrei~A. Rusu, Kieran Milan, John Quan, Tiago Ramalho, Agnieszka
  Grabska-Barwinska, Demis Hassabis, Claudia Clopath, Dharshan Kumaran, and
  Raia Hadsell.
\newblock Overcoming catastrophic forgetting in neural networks.
\newblock {\em Proceedings of the National Academy of Sciences},
  114(13):3521--3526, 2017.

\bibitem{PARISI201954}
German~I. Parisi, Ronald Kemker, Jose~L. Part, Christopher Kanan, and Stefan
  Wermter.
\newblock Continual lifelong learning with neural networks: A review.
\newblock {\em Neural Networks}, 113:54--71, 2019.

\bibitem{evobones}
Branka Hrvoj-Mihic, Thibault Bienvenu, Lisa Stefanacci, Alysson Muotri, and
  Katerina Semendeferi.
\newblock Evolution, development, and plasticity of the human brain: from
  molecules to bones.
\newblock {\em Frontiers in Human Neuroscience}, 7, 2013.

\bibitem{fremaux2016neuromodulated}
Nicolas Fr{\'e}maux and Wulfram Gerstner.
\newblock Neuromodulated spike-timing-dependent plasticity, and theory of
  three-factor learning rules.
\newblock {\em Frontiers in neural circuits}, 9:85, 2016.

\bibitem{gerstner2018eligibility}
Wulfram Gerstner, Marco Lehmann, Vasiliki Liakoni, Dane Corneil, and Johanni
  Brea.
\newblock Eligibility traces and plasticity on behavioral time scales:
  experimental support of neohebbian three-factor learning rules.
\newblock {\em Frontiers in neural circuits}, 12:53, 2018.

\bibitem{bellec2020solution}
Guillaume Bellec, Franz Scherr, Anand Subramoney, Elias Hajek, Darjan Salaj,
  Robert Legenstein, and Wolfgang Maass.
\newblock A solution to the learning dilemma for recurrent networks of spiking
  neurons.
\newblock {\em Nature communications}, 11(1):1--15, 2020.

\bibitem{Schmidgall2021SpikePropamineDP}
Samuel Schmidgall, Julia Ashkanazy, Wallace~E. Lawson, and Joe Hays.
\newblock Spikepropamine: Differentiable plasticity in spiking neural networks.
\newblock {\em Frontiers in Neurorobotics}, 15, 2021.

\bibitem{10.3389/fnbot.2019.00081}
Jacques Kaiser, Michael Hoff, Andreas Konle, J.~Camilo Vasquez~Tieck, David
  Kappel, Daniel Reichard, Anand Subramoney, Robert Legenstein, Arne Roennau,
  Wolfgang Maass, and Rüdiger Dillmann.
\newblock Embodied synaptic plasticity with online reinforcement learning.
\newblock {\em Frontiers in Neurorobotics}, 13, 2019.

\bibitem{doi:10.1073/pnas.2111821118}
Yuhan~Helena Liu, Stephen Smith, Stefan Mihalas, Eric Shea-Brown, and Uygar
  Sümbül.
\newblock Cell-type\&\#x2013;specific neuromodulation guides synaptic credit
  assignment in a spiking neural network.
\newblock {\em Proceedings of the National Academy of Sciences},
  118(51):e2111821118, 2021.

\bibitem{KUSMIERZ2017170}
Łukasz Kuśmierz, Takuya Isomura, and Taro Toyoizumi.
\newblock Learning with three factors: modulating hebbian plasticity with
  errors.
\newblock {\em Current Opinion in Neurobiology}, 46:170--177, 2017.
\newblock Computational Neuroscience.

\bibitem{steinberg2013causal}
Elizabeth~E Steinberg, Ronald Keiflin, Josiah~R Boivin, Ilana~B Witten, Karl
  Deisseroth, and Patricia~H Janak.
\newblock A causal link between prediction errors, dopamine neurons and
  learning.
\newblock {\em Nature neuroscience}, 16(7):966--973, 2013.

\bibitem{schultz1993responses}
Wolfram Schultz, Paul Apicella, and Tomas Ljungberg.
\newblock Responses of monkey dopamine neurons to reward and conditioned
  stimuli during successive steps of learning a delayed response task.
\newblock {\em Journal of neuroscience}, 13(3):900--913, 1993.

\bibitem{seamans2007dopamine}
Jeremy Seamans.
\newblock Dopamine anatomy.
\newblock {\em Scholarpedia}, 2(6):3737, 2007.

\bibitem{zhang2009gain}
Ji-Chuan Zhang, Pak-Ming Lau, and Guo-Qiang Bi.
\newblock Gain in sensitivity and loss in temporal contrast of stdp by
  dopaminergic modulation at hippocampal synapses.
\newblock {\em Proceedings of the National Academy of Sciences},
  106(31):13028--13033, 2009.

\bibitem{Speranza2021DopamineTN}
Luisa Speranza, Umberto di~Porzio, Davide Viggiano, Antonio de~Donato, and
  Floriana Volpicelli.
\newblock Dopamine: The neuromodulator of long-term synaptic plasticity, reward
  and movement control.
\newblock {\em Cells}, 10, 2021.

\bibitem{ranganath2003neural}
Charan Ranganath and Gregor Rainer.
\newblock Neural mechanisms for detecting and remembering novel events.
\newblock {\em Nature Reviews Neuroscience}, 4(3):193--202, 2003.

\bibitem{teles2013cholinergic}
Leonor Teles-Grilo~Ruivo and Jack Mellor.
\newblock Cholinergic modulation of hippocampal network function.
\newblock {\em Frontiers in synaptic neuroscience}, 5:2, 2013.

\bibitem{brzosko2015retroactive}
Zuzanna Brzosko, Wolfram Schultz, and Ole Paulsen.
\newblock Retroactive modulation of spike timing-dependent plasticity by
  dopamine.
\newblock {\em Elife}, 4:e09685, 2015.

\bibitem{hasselmo2006role}
Michael~E Hasselmo.
\newblock The role of acetylcholine in learning and memory.
\newblock {\em Current opinion in neurobiology}, 16(6):710--715, 2006.

\bibitem{zannone2018acetylcholine}
Sara Zannone, Zuzanna Brzosko, Ole Paulsen, and Claudia Clopath.
\newblock Acetylcholine-modulated plasticity in reward-driven navigation: a
  computational study.
\newblock {\em Scientific reports}, 8(1):1--20, 2018.

\bibitem{Zannone2018AcetylcholinemodulatedPI}
Sara Zannone, Zuzanna Brzosko, Ole Paulsen, and Claudia Clopath.
\newblock Acetylcholine-modulated plasticity in reward-driven navigation: a
  computational study.
\newblock {\em Scientific Reports}, 8, 2018.

\bibitem{Hasselmo2006TheRO}
Michael~E. Hasselmo.
\newblock The role of acetylcholine in learning and memory.
\newblock {\em Current Opinion in Neurobiology}, 16:710--715, 2006.

\bibitem{clune2019ai}
Jeff Clune.
\newblock Ai-gas: Ai-generating algorithms, an alternate paradigm for producing
  general artificial intelligence.
\newblock {\em arXiv preprint arXiv:1905.10985}, 2019.

\bibitem{hospedales2020meta}
Timothy Hospedales, Antreas Antoniou, Paul Micaelli, and Amos Storkey.
\newblock Meta-learning in neural networks: A survey.
\newblock {\em arXiv preprint arXiv:2004.05439}, 2020.

\bibitem{finn2017model}
Chelsea Finn, Pieter Abbeel, and Sergey Levine.
\newblock Model-agnostic meta-learning for fast adaptation of deep networks.
\newblock In {\em International conference on machine learning}, pages
  1126--1135. PMLR, 2017.

\bibitem{rothfuss2018promp}
Jonas Rothfuss, Dennis Lee, Ignasi Clavera, Tamim Asfour, and Pieter Abbeel.
\newblock Promp: Proximal meta-policy search.
\newblock {\em arXiv preprint arXiv:1810.06784}, 2018.

\bibitem{fakoor2019meta}
Rasool Fakoor, Pratik Chaudhari, Stefano Soatto, and Alexander~J Smola.
\newblock Meta-q-learning.
\newblock {\em arXiv preprint arXiv:1910.00125}, 2019.

\bibitem{liu2019taming}
Hao Liu, Richard Socher, and Caiming Xiong.
\newblock Taming maml: Efficient unbiased meta-reinforcement learning.
\newblock In {\em International conference on machine learning}, pages
  4061--4071. PMLR, 2019.

\bibitem{metz2018meta}
Luke Metz, Niru Maheswaranathan, Brian Cheung, and Jascha Sohl-Dickstein.
\newblock Meta-learning update rules for unsupervised representation learning.
\newblock {\em arXiv preprint arXiv:1804.00222}, 2018.

\bibitem{andrychowicz2016learning}
Marcin Andrychowicz, Misha Denil, Sergio Gomez, Matthew~W Hoffman, David Pfau,
  Tom Schaul, Brendan Shillingford, and Nando De~Freitas.
\newblock Learning to learn by gradient descent by gradient descent.
\newblock {\em Advances in neural information processing systems}, 29, 2016.

\bibitem{irie2022modern}
Kazuki Irie, Imanol Schlag, R{\'o}bert Csord{\'a}s, and J{\"u}rgen Schmidhuber.
\newblock A modern self-referential weight matrix that learns to modify itself.
\newblock {\em arXiv preprint arXiv:2202.05780}, 2022.

\bibitem{bello2017neural}
Irwan Bello, Barret Zoph, Vijay Vasudevan, and Quoc~V Le.
\newblock Neural optimizer search with reinforcement learning.
\newblock In {\em International Conference on Machine Learning}, pages
  459--468. PMLR, 2017.

\bibitem{zoph2016neural}
Barret Zoph and Quoc~V Le.
\newblock Neural architecture search with reinforcement learning.
\newblock {\em arXiv preprint arXiv:1611.01578}, 2016.

\bibitem{real2019regularized}
Esteban Real, Alok Aggarwal, Yanping Huang, and Quoc~V Le.
\newblock Regularized evolution for image classifier architecture search.
\newblock In {\em Proceedings of the aaai conference on artificial
  intelligence}, volume~33, pages 4780--4789, 2019.

\bibitem{lian2019towards}
Dongze Lian, Yin Zheng, Yintao Xu, Yanxiong Lu, Leyu Lin, Peilin Zhao, Junzhou
  Huang, and Shenghua Gao.
\newblock Towards fast adaptation of neural architectures with meta learning.
\newblock In {\em International Conference on Learning Representations}, 2019.

\bibitem{liu2018darts}
Hanxiao Liu, Karen Simonyan, and Yiming Yang.
\newblock Darts: Differentiable architecture search.
\newblock {\em arXiv preprint arXiv:1806.09055}, 2018.

\bibitem{houthooft2018evolved}
Rein Houthooft, Yuhua Chen, Phillip Isola, Bradly Stadie, Filip Wolski, OpenAI
  Jonathan~Ho, and Pieter Abbeel.
\newblock Evolved policy gradients.
\newblock {\em Advances in Neural Information Processing Systems}, 31, 2018.

\bibitem{co2021evolving}
John~D Co-Reyes, Yingjie Miao, Daiyi Peng, Esteban Real, Sergey Levine, Quoc~V
  Le, Honglak Lee, and Aleksandra Faust.
\newblock Evolving reinforcement learning algorithms.
\newblock {\em arXiv preprint arXiv:2101.03958}, 2021.

\bibitem{bohnstingl2019neuromorphic}
Thomas Bohnstingl, Franz Scherr, Christian Pehle, Karlheinz Meier, and Wolfgang
  Maass.
\newblock Neuromorphic hardware learns to learn.
\newblock {\em Frontiers in neuroscience}, 13:483, 2019.

\bibitem{Schmidgall2020AdaptiveRL}
Samuel Schmidgall.
\newblock Adaptive reinforcement learning through evolving self-modifying
  neural networks.
\newblock {\em Proceedings of the 2020 Genetic and Evolutionary Computation
  Conference Companion}, 2020.

\bibitem{schmidgall2021stable}
Samuel Schmidgall and Joe Hays.
\newblock Stable lifelong learning: Spiking neurons as a solution to
  instability in plastic neural networks, 2021.

\bibitem{jordan2021evolving}
Jakob Jordan, Maximilian Schmidt, Walter Senn, and Mihai~A Petrovici.
\newblock Evolving interpretable plasticity for spiking networks.
\newblock {\em Elife}, 10:e66273, 2021.

\bibitem{elbrecht2020neuroevolution}
Daniel Elbrecht and Catherine Schuman.
\newblock Neuroevolution of spiking neural networks using compositional pattern
  producing networks.
\newblock In {\em International Conference on Neuromorphic Systems 2020}, pages
  1--5, 2020.

\bibitem{kulkarni2021training}
Shruti Kulkarni, Maryam Parsa, J~Parker Mitchell, and Catherine Schuman.
\newblock Training spiking neural networks with synaptic plasticity under
  integer representation.
\newblock In {\em International Conference on Neuromorphic Systems 2021}, pages
  1--7, 2021.

\bibitem{nessler2008hebbian}
Bernhard Nessler, Michael Pfeiffer, and Wolfgang Maass.
\newblock Hebbian learning of bayes optimal decisions.
\newblock {\em Advances in neural information processing systems}, 21, 2008.

\bibitem{shrestha2018slayer}
Sumit~B Shrestha and Garrick Orchard.
\newblock Slayer: Spike layer error reassignment in time.
\newblock {\em Advances in neural information processing systems}, 31, 2018.

\bibitem{neftci2019surrogate}
Emre~O Neftci, Hesham Mostafa, and Friedemann Zenke.
\newblock Surrogate gradient learning in spiking neural networks: Bringing the
  power of gradient-based optimization to spiking neural networks.
\newblock {\em IEEE Signal Processing Magazine}, 36(6):51--63, 2019.

\bibitem{zenke2021remarkable}
Friedemann Zenke and Tim~P Vogels.
\newblock The remarkable robustness of surrogate gradient learning for
  instilling complex function in spiking neural networks.
\newblock {\em Neural Computation}, 33(4):899--925, 2021.

\bibitem{scherr2020one}
Franz Scherr, Christoph St{\"o}ckl, and Wolfgang Maass.
\newblock One-shot learning with spiking neural networks.
\newblock {\em BioRxiv}, 2020.

\bibitem{lett1975long}
Bow~Tong Lett.
\newblock Long delay learning in the t-maze.
\newblock {\em Learning and Motivation}, 6(1):80--90, 1975.

\bibitem{wenk1998assessment}
Gary~L Wenk.
\newblock Assessment of spatial memory using the t maze.
\newblock {\em Current protocols in neuroscience}, 4(1):8--5, 1998.

\bibitem{dudchenko2001animals}
Paul~A Dudchenko.
\newblock How do animals actually solve the t maze?
\newblock {\em Behavioral neuroscience}, 115(4):850, 2001.

\bibitem{deacon2006t}
Robert~MJ Deacon and J~Nicholas~P Rawlins.
\newblock T-maze alternation in the rodent.
\newblock {\em Nature protocols}, 1(1):7--12, 2006.

\bibitem{engelhard2019specialized}
Ben Engelhard, Joel Finkelstein, Julia Cox, Weston Fleming, Hee~Jae Jang,
  Sharon Ornelas, Sue~Ann Koay, Stephan~Y Thiberge, Nathaniel~D Daw, David~W
  Tank, et~al.
\newblock Specialized coding of sensory, motor and cognitive variables in vta
  dopamine neurons.
\newblock {\em Nature}, 570(7762):509--513, 2019.

\bibitem{morcos2016history}
Ari~S Morcos and Christopher~D Harvey.
\newblock History-dependent variability in population dynamics during evidence
  accumulation in cortex.
\newblock {\em Nature neuroscience}, 19(12):1672--1681, 2016.

\bibitem{BREA201661}
Johanni Brea and Wulfram Gerstner.
\newblock Does computational neuroscience need new synaptic learning paradigms?
\newblock {\em Current Opinion in Behavioral Sciences}, 11:61--66, 2016.
\newblock Computational modeling.

\bibitem{delange2021continual}
Matthias Delange, Rahaf Aljundi, Marc Masana, Sarah Parisot, Xu~Jia, Ales
  Leonardis, Greg Slabaugh, and Tinne Tuytelaars.
\newblock A continual learning survey: Defying forgetting in classification
  tasks.
\newblock {\em IEEE Transactions on Pattern Analysis and Machine Intelligence},
  2021.

\bibitem{lake2017building}
Brenden~M Lake, Tomer~D Ullman, Joshua~B Tenenbaum, and Samuel~J Gershman.
\newblock Building machines that learn and think like people.
\newblock {\em Behavioral and brain sciences}, 40, 2017.

\bibitem{lake2015human}
Brenden~M Lake, Ruslan Salakhutdinov, and Joshua~B Tenenbaum.
\newblock Human-level concept learning through probabilistic program induction.
\newblock {\em Science}, 350(6266):1332--1338, 2015.

\bibitem{jose2010comparison}
Shelton Jose and Kumar Gideon~Praveen.
\newblock Comparison between auditory and visual simple reaction times.
\newblock {\em Neuroscience \& Medicine}, 2010, 2010.

\bibitem{stewart2020online}
Kenneth Stewart, Garrick Orchard, Sumit~Bam Shrestha, and Emre Neftci.
\newblock Online few-shot gesture learning on a neuromorphic processor.
\newblock {\em IEEE Journal on Emerging and Selected Topics in Circuits and
  Systems}, 10(4):512--521, 2020.

\bibitem{soltoggio2008evolutionary}
Andrea Soltoggio, John~A Bullinaria, Claudio Mattiussi, Peter D{\"u}rr, and
  Dario Floreano.
\newblock Evolutionary advantages of neuromodulated plasticity in dynamic,
  reward-based scenarios.
\newblock In {\em Proceedings of the 11th international conference on
  artificial life (Alife XI)}, number CONF, pages 569--576. MIT Press, 2008.

\bibitem{miconi2020backpropamine}
Thomas Miconi, Aditya Rawal, Jeff Clune, and Kenneth~O Stanley.
\newblock Backpropamine: training self-modifying neural networks with
  differentiable neuromodulated plasticity.
\newblock {\em arXiv preprint arXiv:2002.10585}, 2020.

\bibitem{risi2012unified}
Sebastian Risi and Kenneth~O Stanley.
\newblock A unified approach to evolving plasticity and neural geometry.
\newblock In {\em The 2012 International Joint Conference on Neural Networks
  (IJCNN)}, pages 1--8. IEEE, 2012.

\bibitem{velez2017diffusion}
Roby Velez and Jeff Clune.
\newblock Diffusion-based neuromodulation can eliminate catastrophic forgetting
  in simple neural networks.
\newblock {\em PloS one}, 12(11):e0187736, 2017.

\bibitem{beaulieu2020learning}
Shawn Beaulieu, Lapo Frati, Thomas Miconi, Joel Lehman, Kenneth~O Stanley, Jeff
  Clune, and Nick Cheney.
\newblock Learning to continually learn.
\newblock {\em arXiv preprint arXiv:2002.09571}, 2020.

\bibitem{ivanov2021increasing}
Vladimir Ivanov and Konstantinos Michmizos.
\newblock Increasing liquid state machine performance with edge-of-chaos
  dynamics organized by astrocyte-modulated plasticity.
\newblock {\em Advances in Neural Information Processing Systems}, 34, 2021.

\bibitem{gordleeva2021modelling}
Susan~Yu Gordleeva, Yulia~A Tsybina, Mikhail~I Krivonosov, Mikhail~V
  Ivanchenko, Alexey~A Zaikin, Victor~B Kazantsev, and Alexander~N Gorban.
\newblock Modelling working memory in spiking neuron network accompanied by
  astrocytes.
\newblock {\em Frontiers in Cellular Neuroscience}, 15:86, 2021.

\bibitem{gaier2019weight}
Adam Gaier and David Ha.
\newblock Weight agnostic neural networks.
\newblock {\em Advances in neural information processing systems}, 32, 2019.

\bibitem{davies2021advancing}
Mike Davies, Andreas Wild, Garrick Orchard, Yulia Sandamirskaya, Gabriel
  A~Fonseca Guerra, Prasad Joshi, Philipp Plank, and Sumedh~R Risbud.
\newblock Advancing neuromorphic computing with loihi: A survey of results and
  outlook.
\newblock {\em Proceedings of the IEEE}, 109(5):911--934, 2021.

\bibitem{davies2018loihi}
Mike Davies, Narayan Srinivasa, Tsung-Han Lin, Gautham Chinya, Yongqiang Cao,
  Sri~Harsha Choday, Georgios Dimou, Prasad Joshi, Nabil Imam, Shweta Jain,
  et~al.
\newblock Loihi: A neuromorphic manycore processor with on-chip learning.
\newblock {\em Ieee Micro}, 38(1):82--99, 2018.

\bibitem{jin2010implementing}
Xin Jin, Alexander Rast, Francesco Galluppi, Sergio Davies, and Steve Furber.
\newblock Implementing spike-timing-dependent plasticity on spinnaker
  neuromorphic hardware.
\newblock In {\em The 2010 International Joint Conference on Neural Networks
  (IJCNN)}, pages 1--8. IEEE, 2010.

\bibitem{rajendran2019low}
Bipin Rajendran, Abu Sebastian, Michael Schmuker, Narayan Srinivasa, and
  Evangelos Eleftheriou.
\newblock Low-power neuromorphic hardware for signal processing applications: A
  review of architectural and system-level design approaches.
\newblock {\em IEEE Signal Processing Magazine}, 36(6):97--110, 2019.

\bibitem{pehle2022brainscales}
Christian Pehle, Sebastian Billaudelle, Benjamin Cramer, Jakob Kaiser,
  Korbinian Schreiber, Yannik Stradmann, Johannes Weis, Aron Leibfried, Eric
  M{\"u}ller, and Johannes Schemmel.
\newblock The brainscales-2 accelerated neuromorphic system with hybrid
  plasticity.
\newblock {\em arXiv preprint arXiv:2201.11063}, 2022.

\bibitem{gerstner1996neuronal}
Wulfram Gerstner, Richard Kempter, J~Leo Van~Hemmen, and Hermann Wagner.
\newblock A neuronal learning rule for sub-millisecond temporal coding.
\newblock {\em Nature}, 383(6595):76--78, 1996.

\bibitem{markram1997regulation}
Henry Markram, Joachim L{\"u}bke, Michael Frotscher, and Bert Sakmann.
\newblock Regulation of synaptic efficacy by coincidence of postsynaptic aps
  and epsps.
\newblock {\em Science}, 275(5297):213--215, 1997.

\bibitem{bi1998synaptic}
Guo-qiang Bi and Mu-ming Poo.
\newblock Synaptic modifications in cultured hippocampal neurons: dependence on
  spike timing, synaptic strength, and postsynaptic cell type.
\newblock {\em Journal of neuroscience}, 18(24):10464--10472, 1998.

\bibitem{sjostrom2001rate}
Per~Jesper Sj{\"o}str{\"o}m, Gina~G Turrigiano, and Sacha~B Nelson.
\newblock Rate, timing, and cooperativity jointly determine cortical synaptic
  plasticity.
\newblock {\em Neuron}, 32(6):1149--1164, 2001.

\bibitem{morrison2008phenomenological}
Abigail Morrison, Markus Diesmann, and Wulfram Gerstner.
\newblock Phenomenological models of synaptic plasticity based on spike timing.
\newblock {\em Biological cybernetics}, 98(6):459--478, 2008.

\bibitem{10.1371/journal.pone.0025339}
Matthieu Gilson and Tomoki Fukai.
\newblock Stability versus neuronal specialization for stdp: Long-tail weight
  distributions solve the dilemma.
\newblock {\em PLOS ONE}, 6(10):1--18, 10 2011.

\bibitem{rubin_stdpstability}
Jonathan Rubin, Daniel Lee, and Haim Sompolinsky.
\newblock Equilibrium properties of temporally asymmetric hebbian plasticity.
\newblock {\em Physical review letters}, 86:364--7, 02 2001.

\bibitem{stablehebbian}
Mark van Rossum, Guoqiang Bi, and G~Turrigiano.
\newblock Stable hebbian learning from spike timing-dependent plasticity.
\newblock {\em The Journal of neuroscience : the official journal of the
  Society for Neuroscience}, 20:8812--21, 01 2001.

\bibitem{PhysRevE.59.4498}
Richard Kempter, Wulfram Gerstner, and J.~Leo van Hemmen.
\newblock Hebbian learning and spiking neurons.
\newblock {\em Phys. Rev. E}, 59:4498--4514, Apr 1999.

\bibitem{song2000competitive}
Sen Song, Kenneth~D Miller, and Larry~F Abbott.
\newblock Competitive hebbian learning through spike-timing-dependent synaptic
  plasticity.
\newblock {\em Nature neuroscience}, 3(9):919--926, 2000.

\bibitem{10.1371/journal.pbio.0030068}
Sen Song, Per~Jesper Sjöström, Markus Reigl, Sacha Nelson, and Dmitri~B
  Chklovskii.
\newblock Highly nonrandom features of synaptic connectivity in local cortical
  circuits.
\newblock {\em PLOS Biology}, 3(3):null, 03 2005.

\bibitem{gutig2003learning}
Robert G{\"u}tig, Ranit Aharonov, Stefan Rotter, and Haim Sompolinsky.
\newblock Learning input correlations through nonlinear temporally asymmetric
  hebbian plasticity.
\newblock {\em Journal of Neuroscience}, 23(9):3697--3714, 2003.

\bibitem{senn2001algorithm}
Walter Senn, Henry Markram, and Misha Tsodyks.
\newblock An algorithm for modifying neurotransmitter release probability based
  on pre-and postsynaptic spike timing.
\newblock {\em Neural computation}, 13(1):35--67, 2001.

\bibitem{gjorgjieva2011triplet}
Julijana Gjorgjieva, Claudia Clopath, Juliette Audet, and Jean-Pascal Pfister.
\newblock A triplet spike-timing--dependent plasticity model generalizes the
  bienenstock--cooper--munro rule to higher-order spatiotemporal correlations.
\newblock {\em Proceedings of the National Academy of Sciences},
  108(48):19383--19388, 2011.

\bibitem{wang2005coactivation}
Huai-Xing Wang, Richard~C Gerkin, David~W Nauen, and Guo-Qiang Bi.
\newblock Coactivation and timing-dependent integration of synaptic
  potentiation and depression.
\newblock {\em Nature neuroscience}, 8(2):187--193, 2005.

\bibitem{pfister2006triplets}
Jean-Pascal Pfister and Wulfram Gerstner.
\newblock Triplets of spikes in a model of spike timing-dependent plasticity.
\newblock {\em Journal of Neuroscience}, 26(38):9673--9682, 2006.

\bibitem{bienenstock1982theory}
Elie~L Bienenstock, Leon~N Cooper, and Paul~W Munro.
\newblock Theory for the development of neuron selectivity: orientation
  specificity and binocular interaction in visual cortex.
\newblock {\em Journal of Neuroscience}, 2(1):32--48, 1982.

\bibitem{salgado2012noradrenergic}
Humberto Salgado, Georg K{\"o}hr, and Mario Trevino.
\newblock Noradrenergic ‘tone’determines dichotomous control of cortical
  spike-timing-dependent plasticity.
\newblock {\em Scientific reports}, 2(1):1--7, 2012.

\bibitem{fremaux2013reinforcement}
Nicolas Fr{\'e}maux, Henning Sprekeler, and Wulfram Gerstner.
\newblock Reinforcement learning using a continuous time actor-critic framework
  with spiking neurons.
\newblock {\em PLoS computational biology}, 9(4):e1003024, 2013.

\bibitem{bellec2019biologically}
Guillaume Bellec, Franz Scherr, Elias Hajek, Darjan Salaj, Robert Legenstein,
  and Wolfgang Maass.
\newblock Biologically inspired alternatives to backpropagation through time
  for learning in recurrent neural nets.
\newblock {\em arXiv preprint arXiv:1901.09049}, 2019.

\bibitem{bing2018end}
Zhenshan Bing, Claus Meschede, Kai Huang, Guang Chen, Florian Rohrbein, Mahmoud
  Akl, and Alois Knoll.
\newblock End to end learning of spiking neural network based on r-stdp for a
  lane keeping vehicle.
\newblock In {\em 2018 IEEE international conference on robotics and automation
  (ICRA)}, pages 4725--4732. IEEE, 2018.

\bibitem{seung2003learning}
H~Sebastian Seung.
\newblock Learning in spiking neural networks by reinforcement of stochastic
  synaptic transmission.
\newblock {\em Neuron}, 40(6):1063--1073, 2003.

\bibitem{MARDER20121}
Eve Marder.
\newblock Neuromodulation of neuronal circuits: Back to the future.
\newblock {\em Neuron}, 76(1):1--11, 2012.

\end{thebibliography}

\end{document}